%% file: main.tex
\documentclass[runningheads]{llncs}
\usepackage{graphicx}
\usepackage{todonotes}
\usepackage{booktabs}
\usepackage{multirow}
\usepackage{amsmath}
\usepackage{hyperref}
\usepackage{subcaption}
\usepackage[export]{adjustbox}
\usepackage{color,ulem}
\usepackage{array}

\usepackage{xcolor}
\usepackage{colortbl}
\definecolor{blue}{HTML}{4472C4}
\definecolor{green}{HTML}{70AD47}
\definecolor{orange}{HTML}{ED7D31}
\definecolor{purple}{HTML}{7030A0}
\definecolor{yellow}{HTML}{FFC000}
\definecolor{teal}{HTML}{00CC99}
\definecolor{pink}{HTML}{CC0099}

\newcommand{\cfbox}[2]{%
    \colorlet{currentcolor}{.}%
    {\color{#1}%
    \fbox{\color{currentcolor}#2}}%
}
\newcommand{\repeatthanks}{\textsuperscript{\thefootnote}}
\begin{document}
% dummy title, do not hesitate to change it
    \title{Reading Order Independent Metrics for Information Extraction in Handwritten Documents}
\titlerunning{Reading Order Independent Metrics for Information Extraction}
% If the paper title is too long for the running head, you can set
% an abbreviated paper title here
%

\author{
    David Villanova-Aparisi\inst{1}\thanks{Authors contributed equally to this work.}\orcidID{0000-0003-2301-6673} 
    \and Solène Tarride\inst{2}\repeatthanks{}\orcidID{0000-0001-6174-9865} 
    \and Carlos-D. Martínez-Hinarejos\inst{1}\orcidID{0000-0002-6139-2891}
    \and Verónica Romero\inst{3}\orcidID{0000-0002-1721-5732 }
    \and Christopher Kermorvant\inst{2}\orcidID{0000-0002-7508-4080}
    \and Moisés Pastor-Gadea\inst{1}
}

\authorrunning{Villanova-Aparisi and Tarride et al.}
% First names are abbreviated in the running head.
% If there are more than two authors, 'et al.' is used.
%
\institute{PRHLT Research Center, Universitat Politècnica de València, Camí de Vera, s/n, València 46021, Spain \and TEKLIA, Paris, France \and Departament d'Informàtica, Universitat de València, València 46010, Spain}
\maketitle
%

% Paper length = up to 15 pages (excluding references and annex)

\begin{abstract}
Information Extraction processes in handwritten documents tend to rely on obtaining an automatic transcription and performing Named Entity Recognition (NER) over such transcription. For this reason, in publicly available datasets, the performance of the systems is usually evaluated with metrics particular to each dataset. Moreover, most of the metrics employed are sensitive to reading order errors. Therefore, they do not reflect the expected final application of the system and introduce biases in more complex documents. In this paper, we propose and publicly release a set of reading order independent metrics tailored to Information Extraction evaluation in handwritten documents. In our experimentation, we perform an in-depth analysis of the behavior of the metrics to recommend what we consider to be the minimal set of metrics to evaluate a task correctly.

\end{abstract}

\section{Introduction}
\input{Contents/1-Introduction}

\section{Related Work}
\label{sec:related_works}
\input{Contents/2-RelatedWork}

\section{Evaluated Metrics}
\label{sec:metrics}
\input{Contents/3-ProposedMetrics}

\section{Experimental Setup}
\label{sec:experimental-setup}
\input{Contents/4-ExperimentalSetup}

\section{Results}
\label{sec:results}
\input{Contents/5-Results}

\section{Conclusion}
\input{Contents/6-Conclusion}

\section*{Acknowledgments}
The authors would like to thank Yoann Schneider for his careful code review.
This work was supported by  Grant PID2020-116813RB-I00 funded by MCIN/AEI/
10.13039/501100011033, by Grant ACIF/2021/436 funded by Generalitat Valenciana and by Grant PID2021-124719OB-I00 funded by MCIN/AEI/10.13039/
501100011033 and by ERDF, EU A way of making Europe.

%\todo[color=orange]{
%    Linhares Pontes, Elvys, et al. (2019) "Impact of OCR quality on named entity linking"
%    Hamdi, Ahmed, et al. (2019) "An analysis of the performance of named entity recognition over OCRed documents"
%    Hamdi, Ahmed, et al. (2020) "Assessing and minimizing the impact of OCR quality on named entity recognition"
%    Van Strien, Daniel, et al. (2020) "Assessing the impact of OCR quality on downstream NLP tasks"
%    Neudecker, Clemens, et al. (2021) "A survey of OCR evaluation tools and metrics"
%    Ehrmann, Maud, et al. (2020) "Extended overview of CLEF HIPE 2020: named entity processing on historical newspapers"
%    Ehrmann, Maud, et al. (2022) "Extended overview of HIPE-2022: Named entity recognition and linking in multilingual historical documents"
%    Ehrmann, Maud, et al. (2023) "Named entity recognition and classification in historical documents: A survey"
%}

\bibliographystyle{splncs04}
\bibliography{main}

\pagebreak
\section{Annex}
\input{Contents/7-Annex}
\end{document}

%% file: Contents/1-Introduction.tex
%Justify relevance of metrics (works in historical text, difficult layout, expected application of the system doesn't care about reading order in IE). VERY IMPORTANT TO CITE DATABASES IN WHICH THAT IS THE CASE

% I think we can present and discuss different document layouts with NER:
%* records: BALSAC (https://arxiv.org/pdf/2304.14044.pdf) / Esposalles
%    * the order in which information is written change (depending on the priest, the time period, the district...)
%    * for large projects (BALSAC), we have few pages completely annotated with reading order, and a lot of pages with only key-value ground truth 
%    * so the evaluation is done once a database is created (BALSAC)
%* tables: POPP, Socface (https://teklia.com/blog/202109-socface/, nothing published yet, maybe for ICDAR if Mélodie & Christopher find the time)
%    * the table layout can change for a given collection (order of columns change, columns disappear / appear over the years)
%    * the ground truth is often in key-value format (e.g. not always in the right order)
%    * once again, evaluation is done once a database is created (SOCFACE)
%* regular full-text pages: HOME NACR (charters), IAM, HOME Alcar
%    * complex layout (multiple columns) can lead to segmentation errors, leading to reading order errors
%    * if we are only interested in Named Entities, then the reading order should not matter
% forms (business documents, medical records, tax form...): FUNSD (https://guillaumejaume.github.io/FUNSD/), Simara
%    * complex layout in key-value setting

Information Extraction (IE) refers to the identification of parts of digital text that contain specific knowledge. In particular, the task of Named Entity Recognition (NER) \cite{mohit2014named,NER-survey} aims to tag parts of the text that contain specific semantic information with their corresponding categories. Traditionally, the approach used to address this task on both handwritten or typeset documents involves a first step of automatic transcription of the document, followed by tagging of the resulting digital text \cite{IAM-NER,MONROC-NERSurvey,abadie2022jointner}. Coupled approaches that solve the task in one step have also seen successful application in recent years \cite{borocs2020comparison,carbonell2018joint,tarride2022comparative} at solving the task, mainly when applied to historical documents. Moreover, there is a growing trend in the scientific community towards adopting full-page end-to-end models for document understanding \cite{dan,donut}. The challenge now is to determine effective evaluation methods for these models to be compared with other approaches, as the currently employed metrics are sensitive to reading order errors \cite{ReadingOrderMatters}.

Researchers in the field of Automatic Text Recognition (ATR) are actively tackling this issue, either focusing on ways to predict text with the correct reading order \cite{LayoutReader,dan}, or by designing reading order independent metrics \cite{CLAUSNER2020Flexible}. In a recent proposal, Vidal et al. \cite{VIDAL2023PAGEASSESSMENT} discuss how reading order challenges stemming from segmentation errors impact conventional text recognition metrics. They propose several metrics to measure the transcription quality independently of the reading order. Unlike traditional ATR metrics, which were traditionally computed on smaller units like words or text lines to mitigate reading order issues, these proposed metrics provide a more comprehensive assessment of ATR on full documents. While there have been advancements in designing metrics for ATR at full pages, a comparable focus in the realm of IE is still in its early stages.

 %\textit{Zhang et al.} \cite{ReadingOrderMatters} were the first to observe the impact of reading order for IE from printed documents. %They find that generic OCR can lead to reading order errors, making it impossible for sequence-labeling methods to accurately predict Named Entities. To overcome this issue, the authors propose a prediction head that reorders predicting paths within a graph of tokens. 

When performing IE with the intent of building a knowledge database, the goal is to identify which Named Entities appear in each document. Therefore, as long as the segmentation errors do not lead to splitting Named Entities, the order in which they are recognized should not matter. Therefore, with this expected application of IE, we claim that metrics should be independent of reading order. In this paper, we present the following contributions:

%Previously we had:
%IE aims to build a knowledge database with the information contained within its documents. \todo[color=orange]{R3 - claim overly simplistic (There are numerous - in fact a majority of - IE tasks where the aim is not to populate a "database". Instead tasks such as relation extraction, coreference resolution, relation extraction, semantic retrieval all depend on the context of the token appearing in the correct sequence.)} As long as the segmentation errors do not compromise the coherence of the Named Entities, the order in which they appear should not matter for the final application of the system\todo[color=orange]{R3 -  What exactly is meant here by "the coherence of the NEs"? State-of-the-art NER systems like sequence tagging LSTMs or transformer models, rely significantly on the token context appearing in the correct sequence order.}. Therefore, we claim that IE metrics should be independent of reading order. In this paper, we present the following contributions:
\begin{itemize}
    \item We introduce a set of metrics that are not influenced by reading order when evaluating the NER process in automatic transcriptions with flat entities (i.e. we do not consider entities that appear inside other entities);
    \item We conduct a thorough analysis of the considered metrics using a model trained on four publicly available datasets and a real-world use case;
    \item Based on our analysis, we recommend the adoption of the proposed OIECER, OIEWER, and OINerval metrics for future benchmarks, identified as the most effective;
    \item We provide an open-source Python package\footnote{\url{https://pypi.org/project/ie-eval/}} including all the metric implementations.
\end{itemize}

%This paper is organized as follows. Section \ref{sec:related_works} gives an overview of metrics designed for information extraction. In Section \ref{sec:metrics}, we introduce new metrics that are independent of the reading order. Section \ref{sec:experimental-setup} provides details about our methodology and the datasets used in this study. Finally, in Section \ref{sec:results}, we present and discuss the results and provide recommendations for future benchmarks.

%% file: Contents/2-RelatedWork.tex
The evaluation of Information Extraction (IE) models from handwritten and printed documents involves a variety of metrics designed to assess the accuracy and efficiency of automatic models. 

In Natural Language Processing tasks such as part-of-speech tagging, where only the tagging itself is evaluated, metrics like Precision, Recall, and F1-score can be computed directly without requiring alignment between ground truth and predicted texts \cite{NER-survey,CLEFHIPE2022,EntityLinking}. However, when working with automatic transcriptions, the ground truth text and predicted text may differ due to text recognition errors. To evaluate the model, then, it is necessary to obtain the best match between the ground truth Named Entities and the hypothesized ones. In this case, metrics introduce dependencies on word positions or reading order depending on the expected application of the system.

%However, when ground truth and predicted text may differ due to text recognition errors, the Named Entities must be matched before evaluating the model. In this case, metrics come with distinct application criteria and use cases, as most of them introduce dependencies on word positions or reading order. 

Yet, these dependencies are not required in IE applications, such as filling a knowledge database. 
Moreover, these dependencies are likely to introduce a bias depending on the segmentation level at which the system is evaluated when using segmentation-free models \cite{VIDAL2023PAGEASSESSMENT}. 

In light of these considerations, this section reviews existing metrics for IE of scanned handwritten or printed documents. Their application criteria, use cases, and limitations are also discussed. 

\subsection{Metrics Based On Word Position Alignment}

% Intro
Information extraction metrics applied to printed documents often rely on word positions to match ground truth and predicted entities based on their Intersection over Union (IoU). 

% Description
Once words are associated, two evaluation approaches can be used. Some researchers evaluate the text recognition task (CER/WER) and the semantic labeling task (Precision, Recall, and F1 scores) separately \cite{toledo2018,IAM-NER}. In this case, the F1 score measures whether the label assigned to Named Entities matches the ground truth label independently of the content of the predicted text. % classification task

% Use cases
The semantic labeling F1 has been used to evaluate IE systems on handwritten datasets in which word positions are available \cite{IAM-NER,toledo2018}, notably IAM \cite{IAM} and Esposalles \cite{Esposalles}. 

% Limitations
These metrics have two main limitations. First, in many cases, word positions are not available to align ground truth and predicted words, especially in handwritten documents where word segmentation is difficult to achieve. Second, these metrics cannot be used to evaluate segmentation-free models working on lines, paragraphs, or entire pages \cite{dan}.

%Use case: when using OCR that provides word positions = printed documents.
%Metrics based on word position alignment:  
%   * F1 (IAM word - Oliver)
%   *  (Donut https://arxiv.org/pdf/2111.15664.pdf)
%   * P/R/F1 Semantic entity labeling is the task of assigning to each semantic entity a label  / field-level F1 score (Donut https://arxiv.org/pdf/2111.15664.pdf, LayoutLM https://dl.acm.org/doi/10.1145/3394486.3403172, BROS https://arxiv.org/abs/2108.04539)

\subsection{Metrics Based On Text Alignment}
% Intro
In scenarios where word positions are unavailable, the alignment between ground truth and predicted entities is usually done using text alignment. In other words, entities are matched according to the order in which they appear and their edit distance from the ground truth entities.

% Description
The IEHHR metric was designed as part of the IEHHR competition\footnote{\url{https://rrc.cvc.uab.es/?ch=10&com=introduction}} \cite{EsposallesCOMPETITION}, in which participants had to automatically extract information from 17th-century marriage records from the Esposalles dataset \cite{Esposalles}. 
The metric considers text recognition and semantic labeling and can operate at different levels: words, text lines, records, or full pages. For each matched word in the marriage record, the score is set to $1-CER$ if the semantic category is correct; otherwise, it is set to 0. The ECER and EWER metrics \cite{VILLANOVA2022EVALUATION} are an extension of the IEHHR metric, with refined substitution costs depending on the transcription quality and semantic tagging. This metric has been used to assess the extraction of Named Entities from historical documents. The Nerval metric \cite{nerval-gitlab} matches ground truth and predicted entities based on their CER and then computes Precision, Recall, and F1 scores. A threshold of 30\% CER is generally used for this metric. 
Finally, a metric based on the Tree Edit Distance (TED) \cite{zhang-TED} was also proposed \cite{donut}. For this metric, first, ground truth and predicted documents are represented as ordered trees. Then, the TED between ground truth and predicted trees is computed before being normalized by the TED between the label and an empty tree. To compute the Tree Error Rate $TER$, a semantic labeling error costs 1, while a text error is computed as the text edit distance between ground truth and predicted leaves \cite{donut}. The accuracy is obtained by taking $min(100, 100- TER)$.

% Use cases
The IEHHR, ECER, EWER, and Nerval metrics have been used to evaluate IE models on various handwritten documents \cite{VILLANOVA2022EVALUATION,MONROC-NERSurvey,key-value,balsac,Simara}, including tables (POPP \cite{POPP}), records (Balsac \cite{balsac}, Esposalles \cite{Esposalles}), index cards (Simara \cite{Simara}), and books (IAM \cite{IAM}, HOME Alcar \cite{HOME-Alcar}). On the other hand, the TED accuracy has been mainly used to evaluate IE models on printed documents \cite{hwang-etal-2021-spatial-ted,donut,table-ted}, such as forms (FUNSD \cite{FUNSD}) and receipts (CORD \cite{CORD}, SROIE \cite{SROIE}). 

% Limits
Although text alignment-based metrics do not introduce a dependency on word position, they do require the system to extract information in a particular order. For structured documents with a fixed layout, this may not be a problem, as the predicted fields can be reordered according to a template. However, for other documents, such as records, the Named Entities included in a document and the order in which they appear can vary. 

\subsection{Alignment-Free Metrics}
% Intro
To the best of our knowledge, only one alignment-free metric has been proposed to evaluate IE models: the field extraction F1.

% Description
This metric \cite{donut,layoutlm,bros} checks whether extracted fields appear in the ground truth. If a single character is missed or added, the metric considers that the field extraction has failed. It is referred to as word-level F1 when considering tagged words \cite{layoutlm}, or entity-level F1 when considering tagged entities \cite{donut}. 

% Use cases
It has been used to evaluate IE models on printed structured documents, such as forms (FUNSD \cite{FUNSD}) and receipts (CORD \cite{CORD}, SROIE \cite{SROIE}).

% Limitations
The field extraction F1 metric has the advantage of not relying on any alignment between the label and prediction. However, it cannot assess the structure of nested entities. Additionally, it does not consider partial overlaps and is very strict for long entity blocks.

\subsection{Discussion}
A first striking observation is the use of different metrics to evaluate IE models, particularly evident in the discrepancy between the metrics used on printed and handwritten datasets. This creates a significant barrier to effectively comparing different models, especially across datasets. 
As models improve, the gap between handwriting recognition and printed text recognition is narrowing, and we see a need to normalize the use of metrics in both communities.

In addition, the dependence of some metrics on alignment, whether based on word position or reading order, creates inherent challenges in the evaluation process. In response to these issues, we advocate adapting existing metrics by eliminating alignment dependencies, paving the way for fairer and more reliable model evaluations. %To further facilitate standardized assessment, we provide an open-source toolbox designed to streamline and unify the implementation of IE metrics. 
We intend to compare a set of metrics and analyze their correlation and, from this study, select those that we consider the most appropriate to normalize evaluation benchmarks.

%% file: Contents/3-ProposedMetrics.tex
From an Information Extraction perspective of the task, having a correct reading order is not necessary. Furthermore, it may introduce a bias since complex documents, such as those that include marginalia \cite{SANCHEZ2019benchmarks}, are more prone to reading order problems. This section briefly introduces the metrics we have considered in our evaluation and the proposed reading order independent metrics we implemented and tested.

\subsection{Order Dependent Metrics}

\subsubsection{Character Error Rate (CER) and Word Error Rate (WER)}
When evaluating automatic transcriptions of any kind, it is necessary to have some criteria to determine their quality. CER and WER \cite{WAGNER1974CER} are widespread metrics that compute the distance between two transcriptions based on the number of edit operations necessary to convert one text into the other.

When dealing with the evaluation of the NER process, CER and WER are not particularly suitable metrics since they cannot incorporate semantics as they consider tags as single tokens in the transcription. In our experiments, we will use CER and WER metrics to assess the quality of the text recognition task performed by our model.

Moreover, we have decomposed the CER and WER computation for each Named Entity category. This way, we can evaluate the difficulty the model faces when transcribing some particular categories. The results of this evaluation are presented in the Annex \ref{sec:results_category}.

\subsubsection{Entity Character Error Rate (ECER) and Entity Word Error Rate (EWER)}
The ECER and EWER metrics were initially presented in \cite{VILLANOVA2022EVALUATION} and are a generalization of the evaluation criteria presented in \cite{EsposallesCOMPETITION}. To compute the distance between two sequences of Named Entities, an edit distance is calculated where the substitution cost depends first on the correctness of the tagging and, finally, on the quality of the transcription of the Named Entity if it was tagged correctly. In the computation of the ECER, the quality of the transcription is assessed by computing the CER. In contrast, in the EWER, the quality of the transcription is evaluated with the WER. For this article, we employ a revised version of the ECER and EWER metrics in which the score is normalized only by the length of the reference and the maximum substitution cost is one. This revised version of the metric is more correlated with the definition of the WER, where the score is normalized only by the number of words of the reference. Moreover, this makes it directly comparable with the reading order independent adaptation proposed in Section \ref{sec:3.2.1}. 

\subsubsection{Soft-aligned Entity Precision, Recall and F1-Score (Nerval) }
The Nerval\cite{nerval-gitlab} method for computing Precision, Recall, and F1 scores is also based on an edit distance. In Nerval, tagged transcriptions are aligned at Named Entity level and a threshold is set for the acceptable character error percentage. In this alignment, True Positives are hypothesized Named Entities that have been correctly tagged and transcribed when compared to their ground truth counterpart. False Positives are hypothesized Named Entities that do not find an acceptable match in the ground truth. False Negatives are ground truth Named Entities for which no acceptable match was found in the hypothesis. Even though, by default, the threshold is set to 30\%, the metric gives interesting information in the extremes. If we allow a 0\% character error, then only perfect matches are allowed. If a 100\% character error is tolerated, the metric solely evaluates the tagging. %Add comment about thresholds

%\subsubsection{mAPCER} => Future work to perhaps consider for benchmark

\subsection{Assignment-based Proposed Reading Order Independent Metrics}
The first set of proposed reading order independent metrics are adaptations of Nerval, ECER, and EWER in which sequentiality constraints have been eliminated when computing the best alignment. More specifically, we will allow one-to-one relations between Named Entities in any given order. Our formulation for the problem shares the notation presented in \cite{MARZAL1993ALIGNMENT} and is inspired by the concepts presented in \cite{VIDAL2023PAGEASSESSMENT}. Given two sequences of Named Entities, $x$ and $y$, we want to search for a non-sequential alignment $A(x,y)$, which is a sequence of pairs of Named Entity indices $(j,k)$ such that $1 \le j \le |x|$, $1 \le k \le |y|$, and for two distinct pairs $(j,k), (j', k') \in A(x,y)$ it must be satisfied that $j \ne j' \wedge k \ne k'$. From the best non-sequential alignment, we can compute the minimum distance $d(x,y)$ between the two sequences by following the following equation:

\begin{equation}
    d(x,y) = \min_{A(x,y)} \sum_{j,k \in A(x,y)} \delta(x_j, y_k)
\end{equation}

Obtaining the best alignment can be reduced to an assignment problem in which each hypothesized Named Entity $Y_k$ can be associated with each ground truth Named Entity $X_j$ with a cost $\delta(x_j, y_k)$. To solve this assignment problem, we employ the Hungarian Algorithm \cite{KUHN1955HUNGARIAN}. Since the algorithm expects square cost matrices and the length of the Named Entity sequences $x$ and $y$ may differ, we resort to padding whichever sequence is smaller with dummy elements $\lambda$ with an assignment cost $\delta(x_j, y_k)$ equal to inserting or deleting a Named Entity. How the cost matrix is computed is what differentiates the proposed metrics.

\label{sec:3.2.1}
\subsubsection{Reading Order Independent Entity Character Error Rate (OIECER) and Entity Word Error Rate (OIEWER)}
In the adaptation of the ECER and EWER metrics \cite{VILLANOVA2022EVALUATION}, the assignment cost $\delta(x_j, y_k)$ is equal to one if one of the Named Entities is a dummy element $\lambda$ or equal to the cost of substituting $x_j$ with $y_k$ if both Named Entities exist. For the cost of the substitution, the maximum values of CER and WER are going to be capped to one. By saturating these metrics, the substitution cost is never larger than one. Assuming that $c(x_j)$ and $c(y_k)$ are the Named Entity categories of $x_j$ and $y_k$ and that $t(x_j)$ and $t(y_k)$ are the transcriptions belonging to each Named Entity, the costs can be computed as follows:

%\todo[color=green]{R1 - In eq. (2) and (3), why don't you normalize CER? I feel this would make the weights of the assignment matrix more homogeneous and may be a key to another remark below.}

\begin{align}
\delta_{\mathrm{ECER}}(x_j, y_k) = 
\begin{cases}
1 \, & \text{if} \, x_j = \lambda \vee y_k = \lambda \\
1 \, & \text{if} \, c(x_j) \neq c(y_k) \\
\min(1,\mathrm{CER}(t(x_j), t(y_k))) \, & \mathrm{otherwise}
\end{cases}
\end{align}

\begin{align}
    \delta_{\mathrm{EWER}}(x_j, y_k) = 
    \begin{cases}
    1 \, & \text{if} \, x_j = \lambda \vee y_k = \lambda \\
    1 \, & \text{if} \, c(x_j) \neq c(y_k) \\
    \min (1,\mathrm{WER}(t(x_j), t(y_k))) \, & \mathrm{otherwise}
    \end{cases}
\end{align}

From the computed cost matrix, we can obtain the best alignment and, consequently, the distance between both Named Entity sequences:

\begin{equation}
    \mathrm{OIECER}(x,y) = \min_{A(x,y)} \sum_{j,k \in A(x,y)} \delta_{\mathrm{ECER}}(x_j, y_k)
\end{equation}

\begin{equation}
    \mathrm{OIEWER}(x,y) = \min_{A(x,y)} \sum_{j,k \in A(x,y)} \delta_{\mathrm{EWER}}(x_j, y_k)
\end{equation}

From this definition of distance, we can extend the computation of the score as a weighted average over each document $s$ belonging to a corpus $C$ by considering the set of all hypothesized Named Entity sequences $Y$ and the ground truth sequences $X$. For a given sample document $s$, its associated hypothesized Named Entity sequence is $Y_s$ and the ground truth sequence is $X_s$. From here, we compute the score for a given corpus $C$ as follows:

%todo[color=green]{R1 - In eq. (6) and (7), you are implicitly using micro averaging over all samples, while macro-averaging (per word, per zone, per document), or per-entity-type metrics may also provide important insights when developing an IE system. Could it be possible to separate the architecture into 3 stages: assignment, scoring and aggregation?}

\begin{equation}
    \mathrm{OIECER}(C) = \frac{\sum_{s \in C}{\mathrm{OIECER}(X_s, Y_s)}}{\sum_{s \in C}{|X_s|}}
\end{equation}

\begin{equation}
    \mathrm{OIEWER}(C) = \frac{\sum_{s \in C}{\mathrm{OIEWER}(X_s, Y_s)}}{\sum_{s \in C}{|X_s|}}
\end{equation}

\subsubsection{Reading Order Independent Soft-aligned Entity Precision, Recall and F1-Score (OINerval)}
In the adaptation of the Nerval computation for Precision, Recall, and F1 scores, the assignment cost $\delta(x_j, y_k)$ is computed from the definition of $\delta_{\mathrm{ECER}}$, also considering a saturated version of the CER between two transcriptions. For a given character error threshold $M$, the computation of the cost is as follows:

%\todo[color=green]{R1 - In eq. (8), the fact you are not jointly optimizing over detection and CER may lead to suboptimal assignments. Is there a reason for that? If you used a normalized CER, would this be possible?}

\begin{align}
\delta_{\mathrm{NERVAL-M}}(x_j, y_k) = 
\begin{cases}
1 \, & \text{if} \, x_j = \lambda \vee y_k = \lambda \\
2 \, & \text{if} \, c(x_j) \neq c(y_k) \\
\mathrm{otherwise} &
\begin{cases}
2 \, & \text{if} \, \min(1, \mathrm{CER}(t(x_j), t(y_k))) > \mathrm{M} \\
0 \, & \text{if} \, \min(1, \mathrm{CER}(t(x_j), t(y_k))) \le \mathrm{M} \\
\end{cases}
\end{cases}
\end{align}
The assignment cost $\delta(x_j, y_k)$ determines the nature of the match. If the cost is 0, the Named Entity must be considered a True Positive. Suppose the cost is two due to category mismatch or excessive character error. In that case, the hypothesized Named Entity $y_k$ must be considered a False Positive, and the ground truth Named Entity $x_j$ as a False Negative. If the cost is one due to $y_k$ being a dummy symbol $\lambda$, then $x_j$ is a False Negative. Finally, if the cost is one due to $x_j$ being a dummy symbol $\lambda$, then $y_k$ must be considered a False Positive.

This principle can be extended to whole corpora in which, for each document, the number of TP, FP, and FN are calculated between the hypothesized sequence of Named Entities and the ground truth. Calculating the micro Precision, Recall, and F1 scores can be done traditionally:

%\todo[color=green]{R1 - I feel a bit uncomfortable going from eq. (8) to (9): do you really need to check the value of the assignment cost matrix to determine whether we have a {true,false}×{positive,negative}? Isn't it possible to directly optimize/compute the FScore? (and to rename OINerval to OIE-Fscore to be homogeneous with OIE-{CER,WER}?)}

\begin{equation}
\label{eq:9}
    P = \frac{TP}{TP + FP} \;\;\;\;\; R = \frac{TP}{TP + FN} \;\;\;\;\; F_1 = 2 \cdot \frac{P \cdot R}{P + R} 
\end{equation}

\subsection{Bag-of-words-based Reading Order Independent Metrics}

Our second set of reading order independent metrics is based on the bag-of-words concept and it is inspired by the proposal of \cite{VIDAL2023PAGEASSESSMENT}. Here, we represent the tagged text using a vector of counts in which, for each tagged token in the vocabulary $v \in V'$, we count the number of appearances of that token in the reference $f_X(v)$ and in the hypothesis $f_Y(v)$. Note that, in our case, we are considering the Named Entity category as a part of the token and removing text that does not belong to any category, resulting in the vocabulary $V'$.

\subsubsection{Bag-of-tagged-words Word Error Rate (btWER)}
The bag-of-tagged-words Word Error Rate (btWER) is a direct adaptation from the bag-of-words Word Error Rate (bWER) presented in \cite{VIDAL2023PAGEASSESSMENT}. Here, we represent the output of the system and the reference text as sets of tagged words. Instead of considering whole transcriptions in our adaptation, we focus on Named Entities by removing words not tagged as belonging to a Named Entity. With this modified vocabulary $V'$ and considering that $|X|$ is the number of tagged words in the reference and $|Y|$ is the number of tagged words in the hypothesis, we can compute the btWER between a reference $X$ and a hypothesis $Y$. For this, we consider the frequency with which each word $v$ appears in the reference, $f_X(v)$, and in the hypothesis, $f_Y(v)$. The resulting equation is as follows:

%\todo[color=green]{R1 - Eq (10) for btWER is confusing (and contains at least a typo). Explanations should be improved to avoid having to read ref. 29 to understand the logic behind this.}
\begin{equation}
    \label{eq:10}
    \text{btWER}(X,Y) = \frac{1}{2\left|X\right|} \left( \left|\left|X\right|-\left|Y\right|\right| \sum_{v \in V'}{\left|f_X(v) - f_Y(v)\right|} \right)
\end{equation}

\subsubsection{Bag-of-tagged-words Precision, Recall and F1-Score}
From this text representation, it is possible to compute Precision, Recall, and F1-score. We can define the computation of the True Positives (TP), False Positives (FP), and False Negatives (FN) as follows:

\begin{align}
\label{eq:11}
\begin{split}
\mathrm{TP}(X,Y) &= \sum_{v \in V'}{\text{min}(f_X(v), f_Y(v))} \\
\mathrm{FP}(X,Y) &= \sum_{v \in V'}{\text{max}(f_Y(v) - f_X(v), 0)} \\
\mathrm{FN}(X,Y) &= \sum_{v \in V'}{\text{max}(f_X(v) - f_Y(v), 0)}
\end{split}
\end{align}

Similarly to the Nerval evaluation, this computation can be extended to a whole corpus if, for each document, we compute the amount of TP, FP, and FN between the hypothesis and the ground truth and sum them. The micro Precision, Recall, and F1 scores can be computed as in Equation \ref{eq:9}.

\subsection{Bag-of-entities Error Rate (beER)}
Another representation level consists of, instead of considering words as the minimal unit of information, taking whole Named Entities as the vocabulary for the vector of counts. With this representation, we can follow the computation for btWER as in Equation \ref{eq:10}, resulting in a bag-of-entities Error Rate (beER). With this approach, a single character error in the transcription of a word of the Named Entity is penalized as a complete error.

\subsection{Bag-of-entities Precision, Recall and F1-Score}
With count vectors for whole Named Entities, Precision, Recall, and F1-Scores can also be computed as in its bag-of-tagged-words counterpart (see Equation \ref{eq:11}). As with the beER, a single character error in a word of the hypothesis affects the whole Named Entity to which it belongs.

%% file: Contents/4-ExperimentalSetup.tex
\subsection{Datasets}

\begin{figure}[p]
    \centering
    \begin{subfigure}[b]{0.49\textwidth}
         \includegraphics[width=\textwidth]{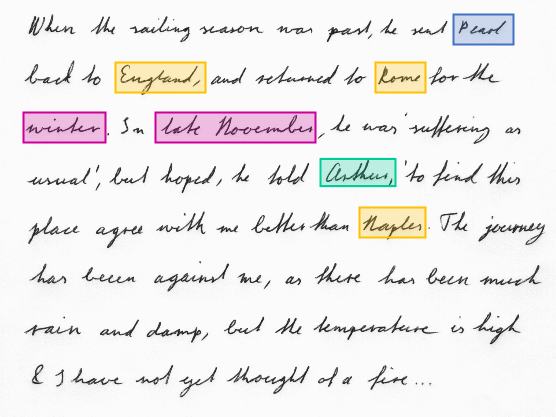}
         \caption{IAM \cite{IAM}. Legend: \cfbox{blue}{\texttt{Object}}; \cfbox{yellow}{\texttt{GPE}}; \cfbox{pink}{\texttt{Date}}; \cfbox{teal}{\texttt{Person}}.\vspace{1em}}
         \label{fig:iam}
     \end{subfigure}
    \begin{subfigure}[b]{0.49\textwidth}
         \includegraphics[width=\textwidth]{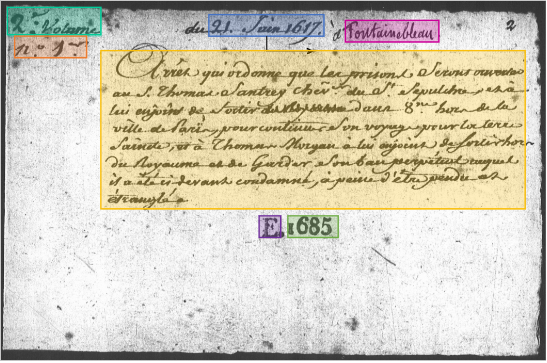}
         \caption{Simara \cite{Simara}. Legend: \cfbox{blue}{\texttt{Date}}; \cfbox{yellow}{\texttt{Title}};  \cfbox{pink}{\texttt{Analysis}}; \cfbox{teal}{\texttt{Arrangement}}; \cfbox{purple}{\texttt{Serie}}; \cfbox{green}{\texttt{Article}}; \cfbox{orange}{\texttt{Volume}}.}
         \label{fig:simara}
     \end{subfigure}
    \begin{subfigure}[b]{0.98\textwidth}
         \includegraphics[width=\textwidth]{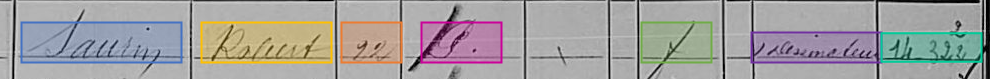}
         \caption{POPP \cite{POPP}. Legend: \cfbox{blue}{\texttt{Surname}}; \cfbox{yellow}{\texttt{Name}}; \cfbox{orange}{\texttt{Birth date}}; \cfbox{pink}{\texttt{Birth location}}; \cfbox{green}{\texttt{Link}}, \cfbox{purple}{\texttt{Occupation}}, \cfbox{teal}{\texttt{Employer}}.}
         \label{fig:popp}
     \end{subfigure}
    \begin{subfigure}[b]{0.98\textwidth}
         \includegraphics[width=\textwidth]{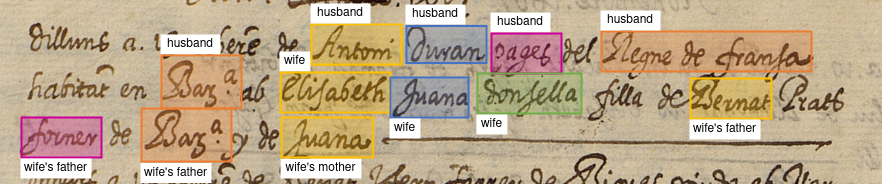}
         \caption{Esposalles \cite{Esposalles}. Legend: \cfbox{yellow}{\texttt{Name}}, \cfbox{blue}{\texttt{Surname}}, \cfbox{pink}{\texttt{Occupation}}, \cfbox{orange}{\texttt{Location}}, \cfbox{green}{\texttt{State}}. Each entity refers to a specific person, as illustrated in the image.}
         \label{fig:esposalles}
     \end{subfigure} 
    \begin{subfigure}[b]{0.98\textwidth}
         \includegraphics[width=\textwidth]{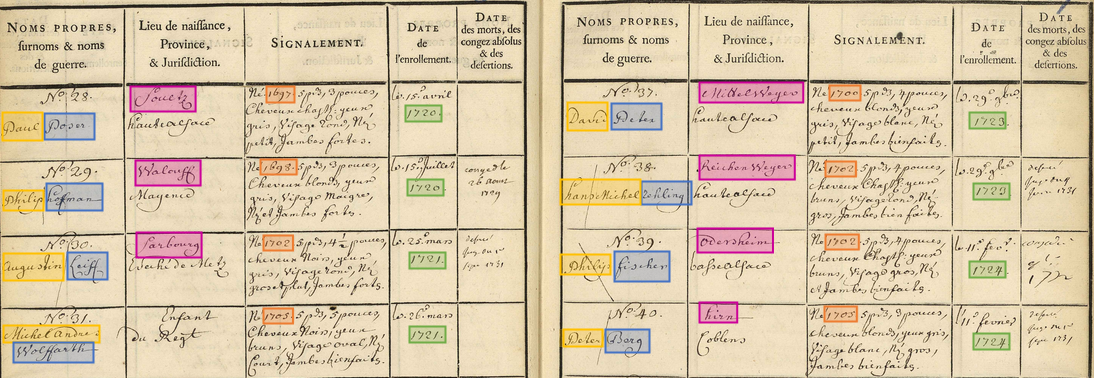}
         \caption{Real use case: military records. Legend: \cfbox{yellow}{\texttt{Name}}, \cfbox{blue}{\texttt{Surname}}, \cfbox{pink}{\texttt{Bith place}}, \cfbox{orange}{\texttt{Birth date}}, \cfbox{green}{\texttt{Date of enrollment}.}}
         \label{fig:military-records}
     \end{subfigure}
     \caption{Illustration of the five datasets used in this study. Named Entities are highlighted. Not all entity types appear in these examples.}
    \label{fig:datasets}
\end{figure}

Five datasets are used in this study: IAM \cite{IAM}, Simara \cite{Simara}, Esposalles \cite{Esposalles}, POPP \cite{POPP}, and a private corpus consisting of French Military Records. An overview of each dataset is presented in Fig. \ref{fig:datasets}

\subsubsection{IAM}
The IAM dataset \cite{IAM} contains modern English documents written by 500 authors. It includes 747 training pages, 116 validation pages, and 336 test pages with their corresponding transcriptions. In 2021, Tüselmann et al. annotated and made available Named Entities in this dataset \cite{IAM-NER}, with 18 tags: \texttt{Cardinal}, \texttt{Date}, \texttt{Event}, \texttt{FAC}, \texttt{GPE}, \texttt{Language}, \texttt{Law}, \texttt{Location}, \texttt{Money}, \texttt{NORP}, \texttt{Ordinal}, \texttt{Organization}, \texttt{Person}, \texttt{Percent}, \texttt{Product}, \texttt{Quantity}, \texttt{Time} and \texttt{Work of art}. 
We use the RWTH split and train a model for the HTR+NER task on paragraphs for our experiments. Note that we do not train on full pages to exclude the printed instructions in the header. 

\subsubsection{Simara}
The Simara dataset \cite{Simara} is a corpus of finding aids from the National Archives of France, dating from the 18th to 20th centuries. It includes 3778 training pages, 811 validation pages, and 804 test pages, with their corresponding information organized in 7 fields: \texttt{Date}, \texttt{Title}, \texttt{Serie}, \texttt{Analysis}, \texttt{Volume\_number}, \texttt{Article\_number}, \texttt{Arrangement}. Documents come from 6 different series where all documents contain the same type of information, but each series has a specific layout. 
We train a model for key-value information extraction on entire pages for our experiments.

\subsubsection{Esposalles}
The ESPOSALLES dataset \cite{Esposalles} is a collection of historical marriage records from the archives of the Cathedral of Barcelona, dating from the 17th century. Each document is written in old Catalan by a single writer, with word, line, and record segmentations. It includes 75 training pages, 25 validation pages, and 25 test pages with their corresponding transcriptions and Named Entities. Each word is also labeled with a semantic category (\texttt{Name}, \texttt{Surname}, \texttt{Occupation}, \texttt{Location}, \texttt{State}, \texttt{Other}) and a person (\texttt{Husband}, \texttt{Wife}, \texttt{Husband’s father}, \texttt{Husband’s mother}, \texttt{Wife’s father}, \texttt{Wife’s mother}, \texttt{Other person}, \texttt{None}). 
We train a model for the HTR+NER task on records for our experiments. Note that the two types of entities are merged (e.g., the tag \texttt{HusbandName} is used for a word labeled as \texttt{Husband} and \texttt{Name}), for a total of 35 combined tags appearing in the dataset.

\subsubsection{POPP}
The POPP dataset \cite{POPP} is a collection of French handwritten tables from the 1926 Paris census. Each document contains 30 rows, and each row describes an individual with 10 columns: \texttt{Surname}, \texttt{Name}, \texttt{Birth date}, \texttt{Birth place}, \texttt{Nationality}, \texttt{Civil status}, \texttt{Link}, \texttt{Education level}, \texttt{Occupation}, and \texttt{Employer}. In our experiments, we use the column name as a Named Entity. It includes 128 training pages, 16 validation pages, and 16 test pages with their corresponding transcriptions and Named Entities.
We train a model for key-value information extraction on text lines for our experiments. 

\subsubsection{Real-World Application: French Military Records}
French Military Records consist of handwritten tables detailing information about soldiers in the French army during the 18th century. Annotations for these records are collected via a crowdsourcing approach, where annotators input information about individual soldiers using a form. However, these annotations are not directly linked to specific areas in the corresponding images, resulting in each two-page document being associated with an unordered (and sometimes incomplete) set of annotations for different soldiers. This annotation collection method is commonly employed in crowdsourcing applications\footnote{\href{https://www.memoiredeshommes.sga.defense.gouv.fr/}{Mémoires des Hommes - French Military Archives}, \href{https://www.archives-nationales.culture.gouv.fr/en/web/guest/home}{Geneanet/Ancestry}, \href{https://www.filae.com/}{Filae}}.
Our model is trained on 118 training and 29 validation full-page images to identify critical information including \texttt{Surname}, \texttt{Name}, \texttt{Birth date}, \texttt{Birth place}, \texttt{Date of enrollment}, while disregarding other information present in the documents.

\subsection{Model}

We train a single DAN model for each dataset for key-value information extraction. 
DAN \cite{dan} is an open-source attention-based Transformer model for handwritten text recognition that can work directly on pages. The encoder is fully convolutional, while the decoder is a Transformer network. It is trained with the cross-entropy loss function. The last layer is a linear layer with a softmax activation function that computes probabilities associated with each vocabulary character. 
The model is trained to recognize characters and unique tokens representing Named Entities. In the case of Simara and POPP, where all words are entities, we use opening tags before each entity, as illustrated in Table \ref{tab:tags+shuffle}. However, since IAM and Esposalles both include words that are not Named Entities, each word is tagged individually with opening tags.

\subsection{Evaluation Methodology}

Once the model is trained, we use it to predict the test set of each dataset. Labels and predictions on the test set are then converted into the IOB2 format\footnote{\url{https://en.wikipedia.org/wiki/Inside-outside-beginning_(tagging)}}. 
Then, the selected metrics are computed on each dataset using a Python package developed for this study. Finally, we shuffle predictions at Named Entity level and re-evaluate them to study the impact of reading order in different metrics. More precisely, we shuffle entity blocks across the document before evaluating with the same metrics. However, the reading order within entity blocks is preserved, as illustrated in Table \ref{tab:tags+shuffle}. 

 \begin{table}[!htb]
    \centering
    \caption{Example of a reference transcription from Simara, and the corresponding shuffled version. Note that we use opening tags to delimit Named Entities.}
    \label{tab:tags+shuffle}
    \vspace{1em}
    \begin{tabular}{p{0.11\textwidth}p{0.88\textwidth}}
    \toprule
         Original & \texttt{\color{yellow}<title>\color{black}AUBERT Huissier priseur à Paris \color{pink}<analysis>\color{black}Contre Baraise \color{blue}<date>\color{black}10 mars 1773 \color{purple}<serie>\color{black}X1A \color{green}<article>\color{black}4723 \color{orange}<reference>\color{black}205} \\
         \midrule
         Shuffled & \texttt{\color{green}<article>\color{black}4723 \color{yellow}<title>\color{black}AUBERT Huissier priseur à Paris \color{purple}<serie>\color{black}X1A \color{orange}<reference>\color{black}205 \color{blue}<date>\color{black}10 mars 1773 \color{pink}<analysis>\color{black}Contre Baraise   } \\
    \bottomrule
    \end{tabular}
\end{table}

To ensure reproducibility of this work, labels and predictions in IOB2 format are publicly available\footnote{\url{https://zenodo.org/records/11083657}}.
%\todo[color=green]{R1 - please host datasets (footnote 7) on some external, durable archival platform.} 
Additionally, we have released an open-source Python package\footnote{\url{https://gitlab.teklia.com/ner/metrics/ie-eval/}} allowing to compute multiple metrics. This package is available on PyPi\footnote{\url{https://pypi.org/project/ie-eval/}}%\todo[color=green]{R1 - code fragmented in 2 or 3 distinct repos, while only one is linked (p. 12): }.

%% file: Contents/5-Results.tex
%Small introduction of the section

Following the experimental method, the model's performance has been evaluated with all the considered metrics in all the available datasets, obtaining the results presented in Table \ref{tab:table5-1}. In this section, we analyze the results and present the main takeaways.
Pearson and Spearman correlation matrix between different metrics across all four datasets are also presented in Annex \ref{sec:correlation}, respectively in Figures \ref{fig:linear-correlation} and \ref{fig:rank-correlation}.

\input{Contents/ALT_T5-RESULTS}

\subsection{Model Analysis Across Datasets}

From Table \ref{tab:table5-1}, we can see the difference in difficulty between the datasets. While the DAN model performs well enough regarding transcription quality across all corpora, it seems almost unable to properly tag Named Entities in the IAM-18 dataset, which was expected since there are few training examples for each entity type. For Simara, POPP, and Esposalles, we consider that the model performs well enough, as it obtains an OINerval-F1 score close to or above 90\% and an OIEWER score close to or below 20\%. 
When comparing correlations across different metrics, we found that DAN showed similar correlation patterns on Simara and POPP. 
On Esposalles, the model exhibits very strong correlations due to nearly flawless text recognition, while on IAM, there is a notable lack of correlation between CER/WER and other metrics due to well-recognized text but poorly identified entities.

\subsection{Impact of Reading Order}
% Key-idea: OD metrics are impacted by shuffling -> not suited for IE
Given the drop in score when using the original Nerval, ECER, and EWER metrics in the shuffled versions of the data, it is clear that they are sensitive to reading order errors. %The only exception to this observation is the ECER and EWER metrics for the IAM database, where the difference falls within the 95\% confidence interval. However, this is due to the already high base error observed for these metrics in the regular data.

% Key-idea: OD and OI metrics yield similar results, but OI metrics have more advantages
Observing the results obtained by the proposed reading-order independent adaptations of Nerval, ECER, and EWER in the case of regular data, we can conclude that they have been well formulated and implemented since the results are practically identical to those of the original metrics. In addition, we observe a perfect correlation between Nerval and OINerval and between ECER/EWER and OIECER/OIEWER metrics in terms of linear and rank correlations. There are slight differences in some corpora, which are to be expected since unconstrained alignments allow a slightly more optimistic evaluation in the case of Named Entity repetition. 

% Key-idea: OI metrics are not impacted by shuffling
The correctness of the implementation is also supported by the fact that OINerval, OIECER, and OIEWER give the same results regardless of whether the data has been shuffled. Given that the proposed versions of the metrics perform as expected while allowing for a more general use case, their use is now recommended when evaluating information extraction tasks.

%Correctness of bag metrics more difficult to verify, but results match our expectations (independent of reading order)
%The bag-of-tagged-words metrics also obtain the same scores independently of the reading order, verifying the implementation. However, since they work at word level instead of considering complete Named Entity blocks, they may not be as useful for evaluating an Information Extraction task. %and may be better considered to evaluate the output of the system as an input to a different process such as Named Entity Linking.

\subsection{Metrics Similarity}
We aim to assess the similarity between order-independent (OI) metrics. Identifying redundancies between metrics is critical to determining which metrics should be used as a reference in future evaluations. To achieve this, we thoroughly analyze the correlations between all the metrics. Correlation matrices are presented in Annex \ref{sec:correlation}. Note that we rely on the absolute correlation for our analysis, as the sign depends on the metric types: error rates (best at 0) or detection rates (best at 100). 

A first observation is that close metrics yield almost perfect correlation: OIECER and OIWER, bt-F1 and btWER, be-F1 and b1-WER. 
Additionally, the CER/WER have relatively low correlations with most IE metrics, as they only evaluate text recognition on all words, not only Named Entities.
Robust rank correlations are observed for all IE metrics except Nerval, suggesting that it provides complementary information about model performance. 
As most other metrics have similar behavior, we recommend using ECER and EWER. These metrics are more appropriate for IE tasks since they operate at the entity level rather than the tagged word level, unlike bt-F1 and btWER, and manage to deal effectively with split errors, unlike be-F1 and beER.
These observations stand for all datasets except for IAM, where the model produces many tagging errors.
%From this analysis, we recommend using ECER, EWER and Nerval in future studies. 

In addition to this quantitative analysis, a qualitative analysis of five prediction scenarios is presented and discussed in Table \ref{tab:qualitative} of the Annex \ref{sec:qualitative}. 

In conclusion,  ECER/EWER and Nerval are complementary and do not correlate strongly, suggesting they provide complementary information about model performance. ECER/EWER evaluate text recognition and labeling errors, while Nerval matches entities using a CER threshold. 
Two scenarios emerge.
If the final information extraction task allows a certain percentage of errors, we recommend using Nerval with the appropriate CER threshold.
However, for a more global evaluation, or if the acceptable percentage of acceptable errors is not known, ECER/EWER provides a more comprehensive evaluation. 

\subsection{Real-World Application: French Military Records}

The French military records dataset presents challenges as the model must extract partial information from complex full-page layouts, compounded by a limited number of training images. While the model was trained on sequentially ordered annotations, allowing it to learn the correct reading order, the test set contains unordered annotations. 
The results on this dataset are shown in Table \ref{tab:table5-1}. The model performs poorly, with only 20\% of the entities being perfectly extracted 
%\todo[color=green]{R1 - is this a typo? This corresponds to the beER score, but the OIE-F1 ("OINerval-F1") is 36\% and seems the correct metric to report here.}
, which is comparable to the performance of DAN on IAM. However, notable improvements are observed in the OIECER and OINerval-F1 metrics compared to IAM due to fewer tagging errors and more text recognition errors. 
It should be noted that order-dependent metrics underestimate model capabilities, primarily due to the misalignment of predictions with the ground truth. In contrast, order-independent (OI) metrics yield more accurate evaluation scores, highlighting the need for OI metrics to improve model evaluation for real-world data with unordered annotations.

%% file: Contents/ALT_T5-RESULTS.tex
%Scores with dd.d format (only 1 decimal)
\setlength\tabcolsep{0.45em}
\begin{table}[htb]
    \centering
    \caption{Comparison between metrics across four datasets. OI refers to Order-Independent metric. Reg. stands for regular data, whereas Shuf. stands for shuffled data. The largest 95\% confidence interval computed is $5.6$ for the Nerval-P in IAM-18 (regular). Therefore, the OI version of the text-alignment metrics obtains a statistically significant difference compared to their original version. }
    \label{tab:table5-1}
    \begin{tabular}{|l|rr|rr|rr|rr|r|}
        \toprule       
        & \multicolumn{2}{c|}{\textbf{IAM-18}} & \multicolumn{2}{c|}{\textbf{Simara}} & \multicolumn{2}{c|}{\textbf{Esposalles}} & \multicolumn{2}{c|}{\textbf{POPP}} & \textbf{Military}\\ 
        \textbf{Metric}  & Reg. & Shuf. & Reg. & Shuf. & Reg. & Shuf. & Reg. & Shuf. & Shuf. \\
        \midrule

        ECER $\downarrow$ &  86.5 &  98.2 & 5.0 & 74.0 & 3.6 & 88.5 & 9.2 & 79.2 & 93.0 \\ 
        OIECER $\downarrow$ &  83.7 &  83.7 & 5.0 & 5.0  & 3.5 &  3.5 &  9.2 & 9.2 & 68.1\vspace{0.4em} \\   
        
        EWER $\downarrow$ &  93.3 &  103.3 & 10.2 & 75.5 & 4.8 & 88.8 & 20.9 & 82.4 & 109.4\\ 
        OIEWER $\downarrow$ &  91.5 &  91.5 & 10.2 & 10.2 &  4.7 & 4.7 & 20.9 & 20.9 & 95.1\vspace{0.4em}\\ 
        
        Nerval-P $\uparrow$ & 29.0 & 2.9 & 95.0 & 37.7 & 96.3 & 17.0 & 89.3 & 35.0 & 14.3 \\
        OINerval-P $\uparrow$ & 29.0 &  29.0 & 95.0 & 95.0 & 96.3 & 96.3 & 89.3 & 89.3 & 33.9\vspace{0.4em} \\ 
        
        Nerval-R $\uparrow$ & 26.6 & 2.7 & 95.3 & 37.9 & 97.1 & 17.2 & 90.0 & 35.3 & 15.1 \\
        OINerval-R $\uparrow$ &  26.6 &  26.6 & 95.4 & 95.4 & 97.1 & 97.1 & 90.0 & 90.0 & 36.0\vspace{0.4em}  \\ 
        
        Nerval-F1 $\uparrow$ & 27.7 & 2.8 & 95.1 & 37.8 & 96.7 & 17.1 & 89.6 & 35.1 & 14.7 \\
        OINerval-F1 $\uparrow$ &  27.7 &  27.7 &  95.2 & 95.2 & 96.7 & 96.7 & 89.6 & 89.6 & 34.9\vspace{0.4em} \\
        
        %BoTWER $\downarrow$ & 100.00 & 100.00 & 31.28 & 31.28 & 7.96 & 7.96 & 44.86 & 44.86 \\ % equation 7
        btWER $\downarrow$ & 85.3 & 85.3 & 17.4 & 17.4 & 4.6 & 4.6 & 24.2 & 24.2 & 92.2\vspace{0.4em} \\  % equation 8
    
        bt-P $\uparrow$ & 29.2 & 29.2 & 84.8 & 84.8 & 95.9  & 95.9 & 77.4 & 77.4 & 21.2 \\
        bt-R $\uparrow$ & 21.2 & 21.2 & 83.7 & 83.7 & 96.2 & 96.2 & 78.0 & 78.0 & 20.9 \\ 
        bt-F1 $\uparrow$ & 24.5 & 24.5 & 84.3 & 84.3 & 96.0 & 96.0 & 77.7 & 77.7 & 21.0\vspace{0.4em}\\ 
        
        beER $\downarrow$ & 94.2 &  94.2 & 23.2 & 23.2 & 5.8 & 5.8 & 23.1 & 23.1 & 96.2\vspace{0.4em} \\ %entity based = field-based F1 from donut -> to check
        
        be-P $\uparrow$ & 21.6 & 21.6 & 77.1 & 77.1 & 94.4 & 94.4 & 77.4 & 77.4 & 18.8 \\
        be-R $\uparrow$ & 19.8 & 19.8 & 77.3 & 77.3 & 95.2 & 95.2 & 78.0 & 78.0 & 20.0 \\ 
        be-F1 $\uparrow$ & 20.7 & 20.7 & 77.2  & 77.2 & 94.8 & 94.8 & 77.7 & 77.7  & 19.4 \\ 
        \midrule
        CER $\downarrow$ & 5.0 & 74.2 & 6.4 & 39.7 & 0.4 & 68.1 & 8.8 & 67.5 & 80.8 	\\ 
        WER $\downarrow$ & 15.5 & 94.0 & 17.3 & 54.2 & 1.5 & 84.2 & 23.6 & 80.5 & 106.1\vspace{0.4em}\\ 
        %entity based = field-based F1 from donut

   %     TED-Acc $\uparrow$ & 18.7 & 16.0 & 92.7 & 92.7 & 95.7 & 93.6 &  90.3 & 90.3 \\ %TED-acc from donut
        \bottomrule
    \end{tabular}
\end{table}

%% file: Contents/6-Conclusion.tex
%\todo[color=orange]{R3 - More complex NER tagging cases such as NEs with (multiple levels of) embedded NEs are not discussed.}

%\todo[color=red]{R2 - what is really necessary is to verify the usefulness of the proposed metric.}

In this article, we have proposed reading-order-independent versions of the Nerval, ECER, and EWER metrics and the adaptation of two bag-of-words-based evaluation schemes for NER in scanned documents. The behavior of the considered metrics has been evaluated in four publicly available corpora, proving that our proposal mimics the original behavior of the metrics while remaining robust to reading order errors. We have analyzed the correlation between the proposed metrics and, considering this study and the expected application of the system, we have chosen ECER and EWER as the general use case metrics to employ from now on. However, if the threshold of tolerable character error is known, we recommend the usage of Nerval. In future work, we would like to extend the usage of these metrics to models that work at different segmentation levels (i.e., page level, line level, and word level), providing a benchmark for each approach. We also plan to extend these metrics to handle nested entities.

%\todo[color=green]{R1 - In your discussion about existing approaches, you point the need for dealing with nested entities, and appropriately mention the tree edit distance used in the DONUT paper. However, you did not implement/compare this metric, and you "forget" in the rest of the paper about nested (I'd say "structured with belongs-to" relations) entities. I know we lack public datasets for this, but I was a bit disappointed as I first thought you dealt with them. Maybe stating this explicitly earlier would be better.}

%We have to choose a couple of metrics as "favorites": OI-EWER, OI-Nerval-F1 (maybe BoTW metric)
%If we do not like BoTW metric, justify it by saying that working at NE blocks imposes a soft alignment, making the metrics almost as strict as original version, whereas when we consider words metrics end up being optimistic
%As future work, we can mention the application of these metrics to an experiment in which different models work at different segmentation levels, providing a benchmark for every model but with the same metric.

%% file: Contents/7-Annex.tex
\subsection{Detailed results by category}
\label{sec:results_category}

We present the details of most metrics for each category on Esposalles in Table \ref{tab:category_results_esposalles}, on IAM in Table \ref{tab:category_results_iam}, on Simara in Table \ref{tab:category_results_simara}, and on POPP in Table \ref{tab:category_results_popp}.

%\todo[color=green]{R1 - for Annex 7.1, table 3-6 should report the number of samples for each category}

\begin{table}[ht]
    \centering
    \caption{Metrics for each Named Entity category on the test set of the Esposalles database.}
    \label{tab:category_results_esposalles}
    \begin{tabular}{l|cc|r}
        \toprule
        \textbf{Category} & \textbf{CER $\downarrow$} & \textbf{WER $\downarrow$} & Number of entities \\ % & \textbf{BoW F1 $\uparrow$} & \textbf{BoTW F1 $\uparrow$} & \textbf{BoE F1 $\uparrow$} & \textbf{Nerval F1 $\uparrow$}\\
        \midrule
        husband location & 3.1 & 5.4 & 250 \\ %& 96.2 & 96.2 & 92.0 & 96.8 \\
        husband name & 3.7 & 4.8 & 253 \\ %& 96.4 & 96.4 & 95.5 & 97.4 \\
        husband occupation & 4.7 & 7.0 & 250 \\ %& 95.0 & 95.0 & 93.3 & 95.8 \\
        husband state & 10.0 & 8.5 & 65 \\ %& 95.6 & 95.6 & 95.6 & 95.6 \\
        husband surname & 5.4 & 6.6 & 253 \\ %& 95.5 & 95.5 & 93.9 & 95.5 \\
        husband's father location & 0.0 & 0.0 & 12 \\ % & 100.0 & 100.0 & 100.0 & 100.0\\
        husband's father name & 3.4 & 4.1 & 149 \\ %& 97.6 & 97.6 & 96.0 & 96.0 \\
        husband's father occupation & 3.0 & 3.6 & 143 \\ %& 97.9 & 97.9 & 96.5 & 97.6 \\
        husband's father surname & 5.7 & 7.4 & 143\\ %& 95.6 & 95.6 & 92.8 & 92.9 \\
        husband's mother name & 1.0 & 2.0 & 149 \\ % & 98.4 & 98.4 & 98.00 & 99.0 \\
        husband's mother surname & 100.0 & 100.0 & 1 \\ % & 0.0 & 0.0 & 0.0 & 0.0 \\
        other person location & 100.0 & 100.0 & 2 \\ %& 0.0 & 0.0 & 0.0 & 0.0 \\
        other person name & 2.1 & 2.9 & 66 \\ % & 97.8 & 97.8 & 97.0 & 96.2 \\
        other person state & 100.0 & 100.0 & 1 \\ %& 0.0 & 0.0 & 0.0 & 0.0 \\
        other person surname & 4.0 & 7.3 & 65 \\ %& 94.1 & 94.1 & 92.3 & 96.2 \\
        wife location & 13.1 & 15.2 & 80 \\ %& 91.1 & 91.1 & 86.6 & 86.8 \\
        wife name & 0.8 & 2.5 & 253 \\ %& 97.65 & 97.7 & 98.0 & 99.2\\
        wife occupation & 2.0 & 1.2 & 62 \\ %& 99.4 & 99.41 & 98.39 & 98.4 \\
        wife state & 3.0 & 4.3 & 248 \\ %& 97.6 & 97.6 & 97.6 & 96.8 \\
        wife surname & 43.5 & 50.0 & 3 \\ %& 66.7 & 66.7 & 80.0 & 80.0 \\
        wife's father location & 1.6 & 3.1 & 148 \\ %& 97.7 & 97.7 & 95.4 & 96.8 \\
        wife's father name & 4.4 & 5.2 & 186 \\ %& 96.3 & 96.3 & 97.0 & 97.6 \\
        wife's father occupation & 2.7 & 4.0 & 179 \\ %& 97.3 & 97.3 & 95.8 & 97.5 \\
        wife's father surname & 4.05 & 7.0 & 183 \\ %& 94.4 & 94.4 & 93.9 & 97.1\\
        wife's mother name & 0.2 & 1.1 & 184 \\ %& 99.2 & 99.2 & 99.2 & 99.7 \\
        \textbf{total (including untagged words)} & 0.4 & 1.5 & - \\ %& 97.4 & 96.0 & 94.8 & 96.7 \\
        \bottomrule
    \end{tabular}
\end{table}

\begin{table}[ht]
    \centering
    \caption{Metrics for each Named Entity category on the IAM database.}
    \label{tab:category_results_iam}
    \begin{tabular}{l|cc|r}
        \toprule
        \textbf{Category} & \textbf{CER $\downarrow$} & \textbf{WER $\downarrow$}  & Number of entities   \\ %& \textbf{BoW F1 $\uparrow$} & \textbf{BoTW F1 $\uparrow$} & \textbf{BoE F1 $\uparrow$} & \textbf{Nerval F1 $\uparrow$}  \\
        \midrule
        cardinal & 66.0 & 70.0 & 70 \\ %& 40.4 & 40.4 & 40.0 & 35.7 \\ 
        date & 72.6 & 73.5 & 59 \\ %& 41.6 & 41.6 & 23.6 & 25.2\\ 
        event & 98.9 & 100.0 & 7 \\ %&0.0 & 0.0 & 0.0 & 0.0 \\ 
        fac & 96.2 & 97.2 & 28 \\ %& 5.3 & 5.3 & 0.0 & 0.0 \\ 
        gpe & 64.3 & 78.1 & 74 \\ %& 30.5 & 30.5 & 28.6 & 26.1  \\ 
        language & 100.0 & 100.0 & 4 \\ %& 0.0 & 0.0 & 0.0 & 0.0 \\ 
        location & 94.6 & 95.1 & 28 \\ %& 13.0 & 13.0 & 10.0 & 6.7 \\ 
        money & 100.0 & 100.0 & 2\\ %& 0.0 & 0.0 & 0.0 & 0.0 \\ 
        norp & 74.3 & 79.8 & 44 \\ %& 30.9 & 30.9 & 31.5 & 31.0 \\ 
        ordinal & 37.6 & 39.5 & 31 \\ %& 71.9 & 71.9 & 71.9 & 74.6 \\ 
        org & 66.9 & 77.7 & 35 \\ %& 31.7 & 31.7 & 21.6 & 14.1 \\ 
        per & 63.8 & 76.9 & 211 \\%& 32.9 & 32.9 & 26.8 & 36.6 \\ 
        percent & 68.2 & 75.0 & 1 \\%& 33.3 & 33.3 & 0.0 & 0.0 \\ 
        product & 67.9 & 72.0 & 13 \\%& 40.0 & 40.0 & 8.7 & 11.8\\ 
        quantity & 97.4 & 98.5 & 19 \\% & 2.9 & 2.9 & 0.0 & 0.0 \\ 
        time & 97.8 & 98.3 & 42 \\% & 3.4 & 3.4 & 3.7 & 3.6 \\ 
        work of art & 93.2 & 97.3 & 38 \\%& 4.9 & 4.9 & 0.0 & 0.0 \\ 
        \textbf{total (including untagged words)} & 5.0 & 15.5 & - \\%& 44.4 & 24.5 & 20.7 & 27.7 \\ 
        \bottomrule
    \end{tabular}
\end{table}

\begin{table}[ht]
    \centering
    \caption{Metrics for each Named Entity category on the Simara database.}
    \label{tab:category_results_simara}
    \begin{tabular}{l|cc|r}
        \toprule
        \textbf{Category} & \textbf{CER $\downarrow$} & \textbf{WER $\downarrow$} & Number of entities \\ % & \textbf{BoW F1 $\uparrow$} & \textbf{BoTW F1 $\uparrow$} & \textbf{BoE F1 $\uparrow$} & \textbf{Nerval F1 $\uparrow$} \\
        \midrule
        analysis & 9.0 & 22.9 & 771 \\ %& 80.2 & 80.2 & 50.78 & 92.5 \\
        arrangement & 6.3 & 13.6 & 77 \\ % & 86.6 & 86.6 & 74.03 & 93.6 \\
        article & 3.1 & 4.2 & 676 \\ %& 96.8 & 96.8 & 97.49 & 98.3 \\
        serie & 3.1 & 3.2 & 676 \\ %& 97.5 & 97.5 & 97.71 & 97.6 \\
        date & 1.4 & 2.7 & 751 \\ %& 97.5 & 97.5 & 94.46 & 97.8 \\
        title & 7.7 & 21.0 & 804 \\ %& 81.7 & 81.7 & 43.91 & 94.3  \\
        volume & 12.4 & 14.1 & 675 \\ % & 89.1 & 89.1 & 88.21 & 90.7 \\
        \textbf{total (including untagged words)} & 6.4 & 17.3 & - \\ %& 84.9 & 84.3 & 77.20 & 95.1 \\
        \bottomrule
    \end{tabular}
\end{table}

\begin{table}[ht]
    \centering
    \caption{Metrics for each Named Entity category on the POPP database.}
    \label{tab:category_results_popp}
    \begin{tabular}{l|cc|r}
        \toprule
        \textbf{Category} & \textbf{CER $\downarrow$} & \textbf{WER $\downarrow$}  & Number of entities \\% & \textbf{BoW F1 $\uparrow$} & \textbf{BoTW F1 $\uparrow$} & \textbf{BoE F1 $\uparrow$} & \textbf{Nerval F1 $\uparrow$}   \\
        \midrule
        birth date & 4.7 & 18.4 & 466 \\ %& 91.5 & 91.5 & 91.4 & 92.8 \\
        civil status & 14.4 & 15.7 & 225\\ %& 91.8 & 91.8 & 91.8 & 91.4 \\
        employer & 14.5 & 34.5 & 396 \\ %& 69.6 & 69.6 & 70.1 & 88.1 \\
        firstname & 4.7 & 18.4 & 477 \\ %& 81.6 & 81.6 & 81.6 & 94.9 \\
        link & 14.4 & 15.7 & 408 \\ %& 85.7 & 85.7 & 84.9 & 84.9 \\
        location of birth & 15.2 & 25.9 & 443 \\ %& 75.7 & 75.7 & 76.5 & 86.9 \\
        occupation & 14.3 & 36.1 & 331 \\ %& 65.7 & 65.7 & 55.8 & 84.2 \\
        surname & 6.9 & 28.2 & 369 \\ %& 71.9 & 71.9 & 71.8 & 92.8 \\
        \textbf{total (including untagged words)} & 8.8 & 23.6 & - \\ %& 77.8 & 77.7 & 77.7 & 89.6 \\
        \bottomrule
    \end{tabular}
\end{table}

\pagebreak

\subsection{Correlation between the different metrics}
\label{sec:correlation}

\begin{figure}[ht]    \centering
    \begin{subfigure}[b]{0.49\textwidth}
         \includegraphics[width=\textwidth, center]{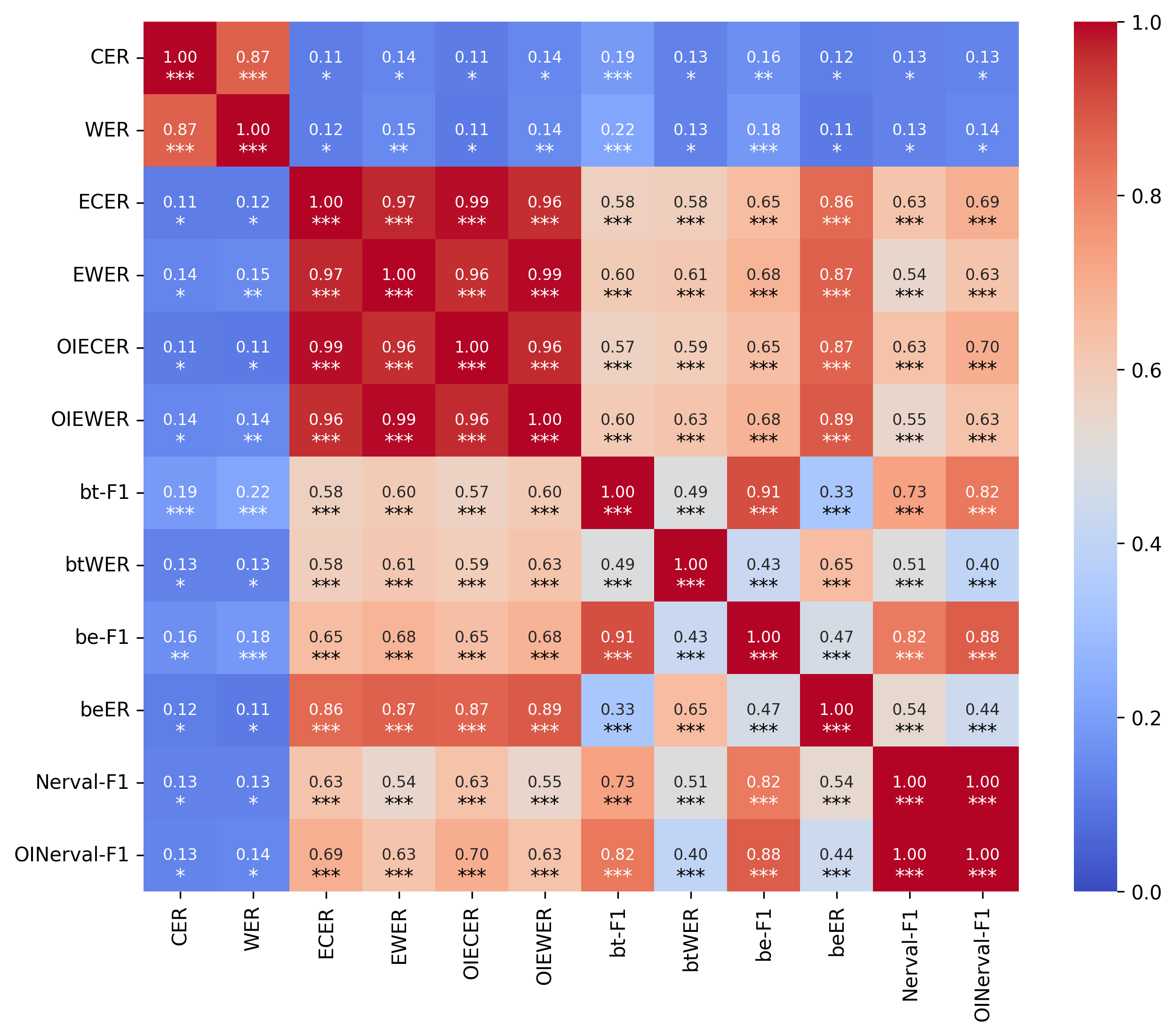}
         \caption{IAM \cite{IAM-NER}}
         \label{fig:iam_l_corr}
     \end{subfigure}   
    \begin{subfigure}[b]{0.49\textwidth}
         \includegraphics[width=\textwidth, center]{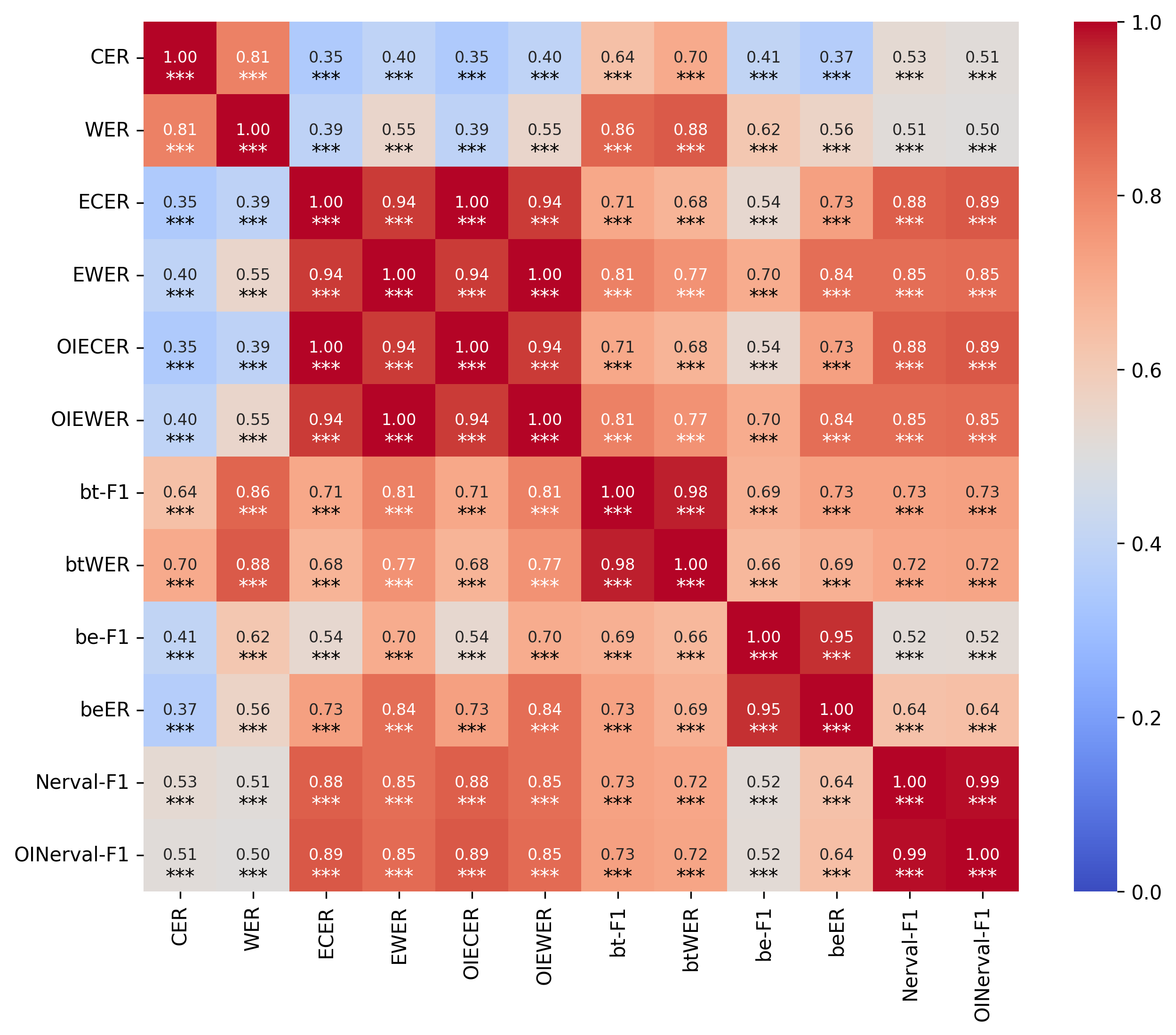}
         \caption{Simara \cite{Simara}}
         \label{fig:simara_l_corr}
     \end{subfigure} 
    \begin{subfigure}[b]{0.49\textwidth}
         \includegraphics[width=\textwidth, center]{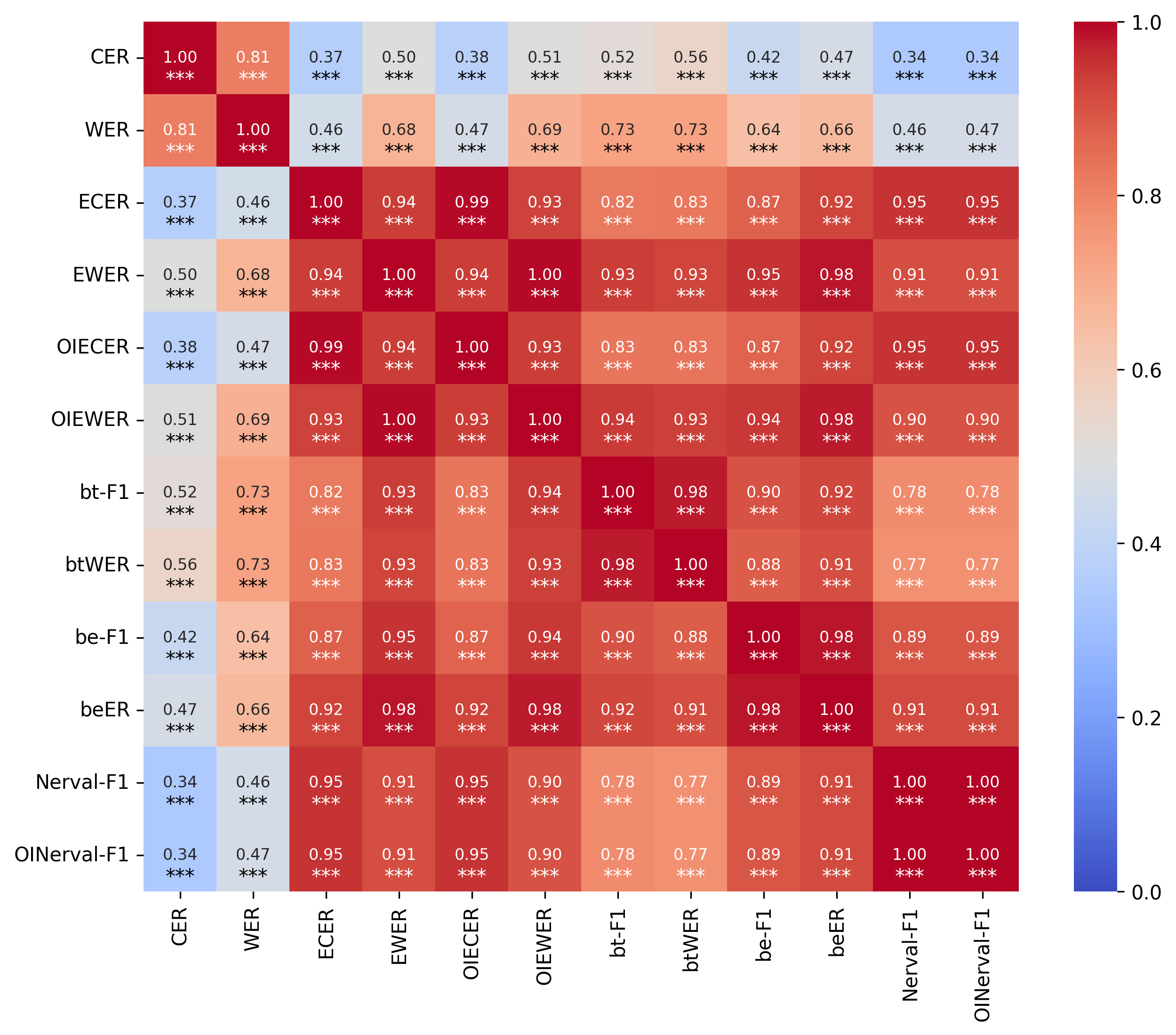}
         \caption{Esposalles \cite{Esposalles}}
         \label{fig:esposalles_l_corr}
     \end{subfigure}   
    \begin{subfigure}[b]{0.49\textwidth}
         \includegraphics[width=\textwidth, center]{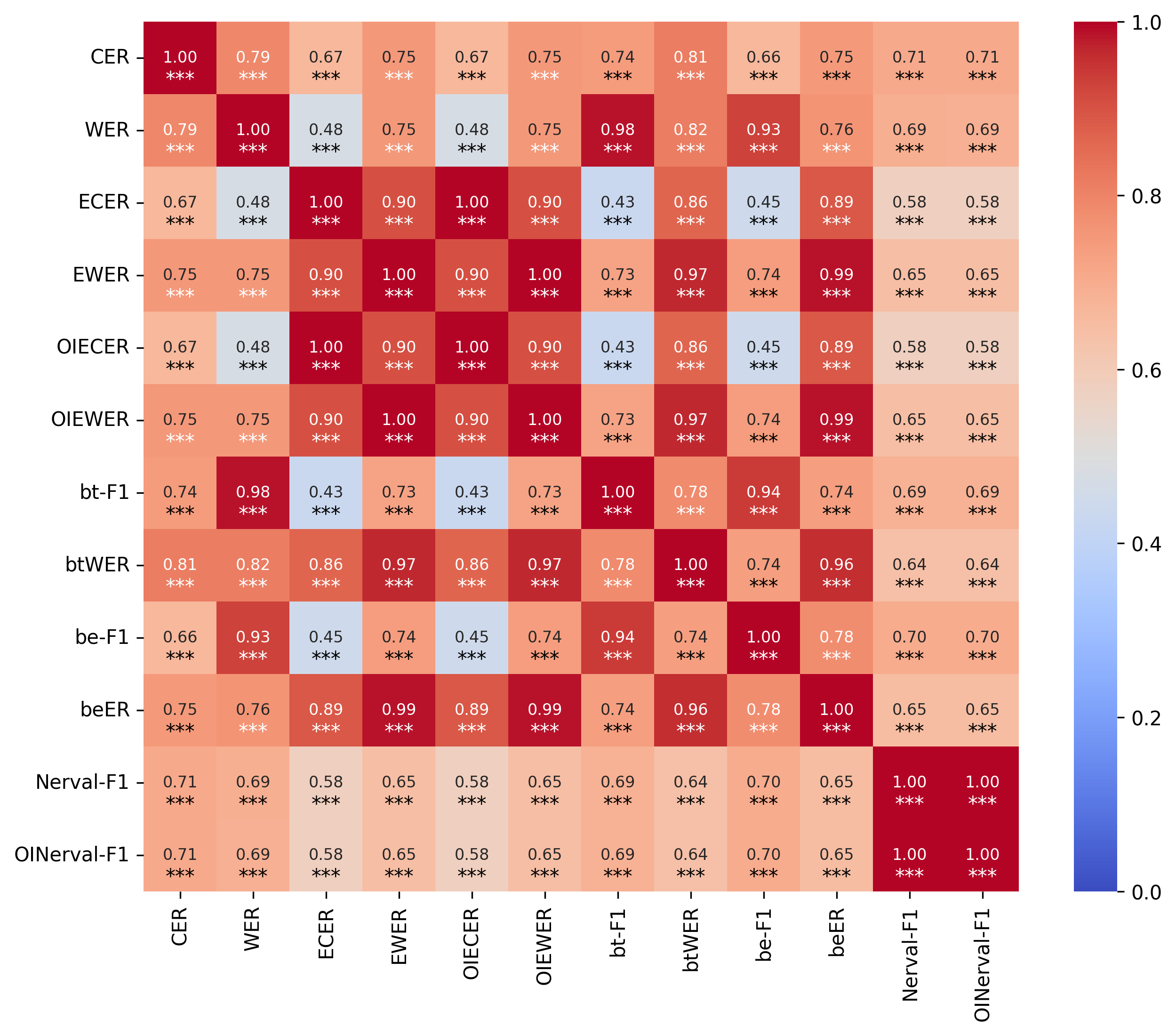}
         \caption{POPP \cite{POPP}}
         \label{fig:popp_l_corr}
     \end{subfigure}   
     \caption{Absolute linear correlation (Pearson) between the different metrics across the four datasets. The correlation value appears on each cell, as well as an indication of its p-value: \texttt{*} indicates a $p-value < 0.05$, \texttt{**} indicates a $p-value < 0.01$, and \texttt{***} indicates a $p-value < 0.001$. No star indicates that the correlation is not significant.}
    \label{fig:linear-correlation}
\end{figure}

\begin{figure}[p]
    \centering
    \begin{subfigure}[b]{0.49\textwidth}
         \includegraphics[width=\textwidth, center]{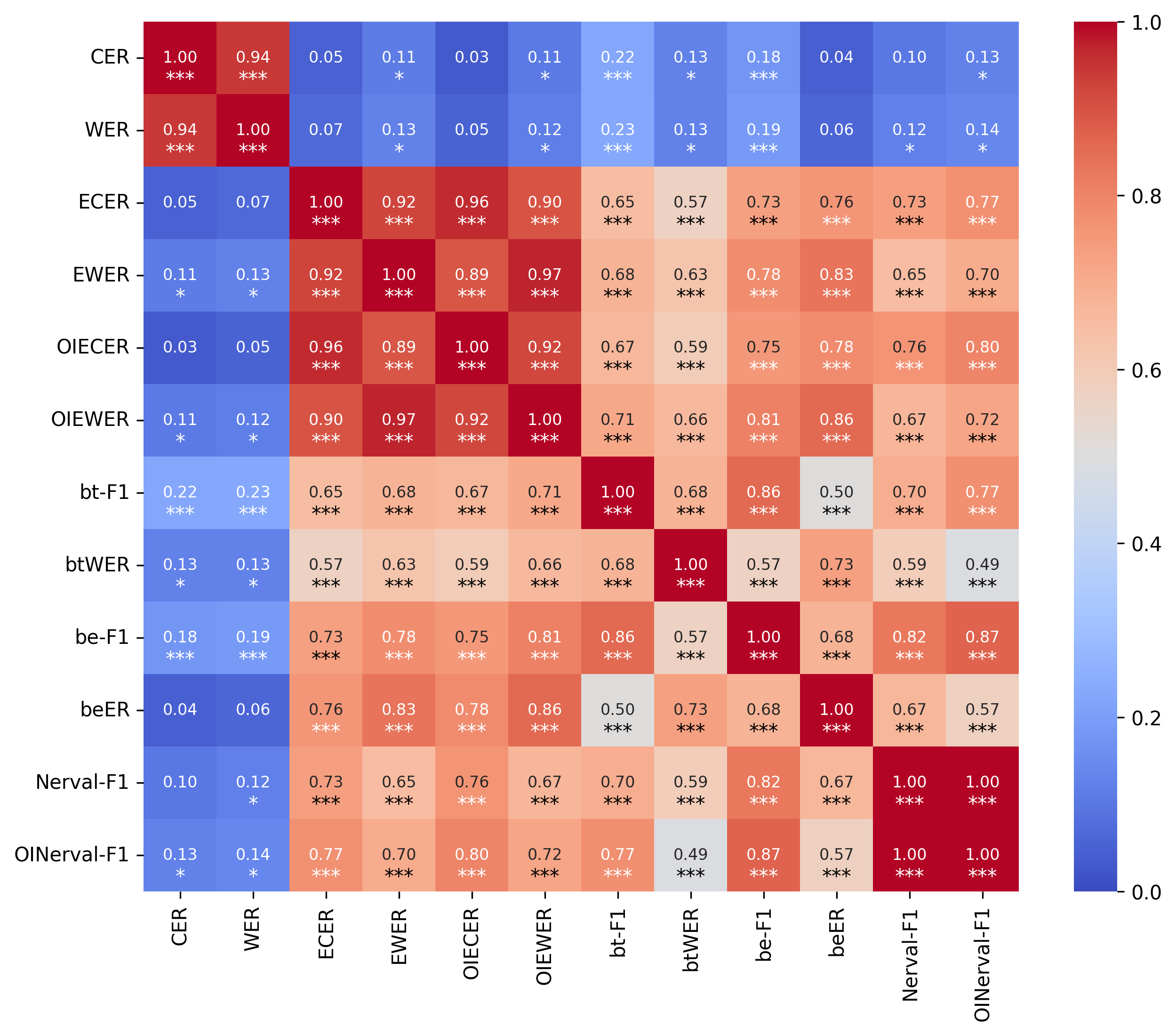}
         \caption{IAM \cite{IAM-NER}}
         \label{fig:iam_s_corr}
     \end{subfigure}   
    \begin{subfigure}[b]{0.49\textwidth}
         \includegraphics[width=\textwidth, center]{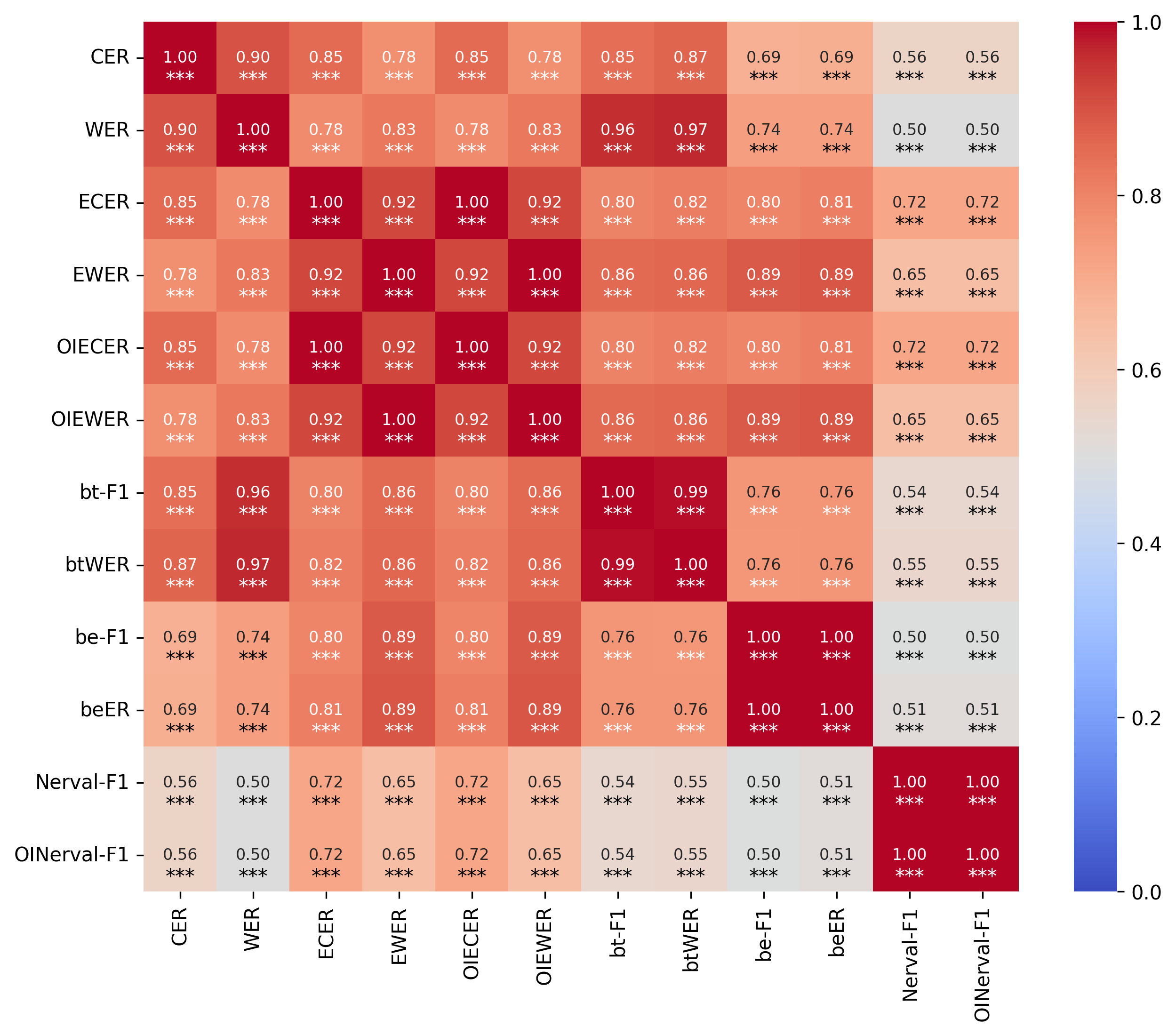}
         \caption{Simara \cite{Simara}}
         \label{fig:simara_s_corr}
     \end{subfigure} 
    \begin{subfigure}[b]{0.49\textwidth}
         \includegraphics[width=\textwidth, center]{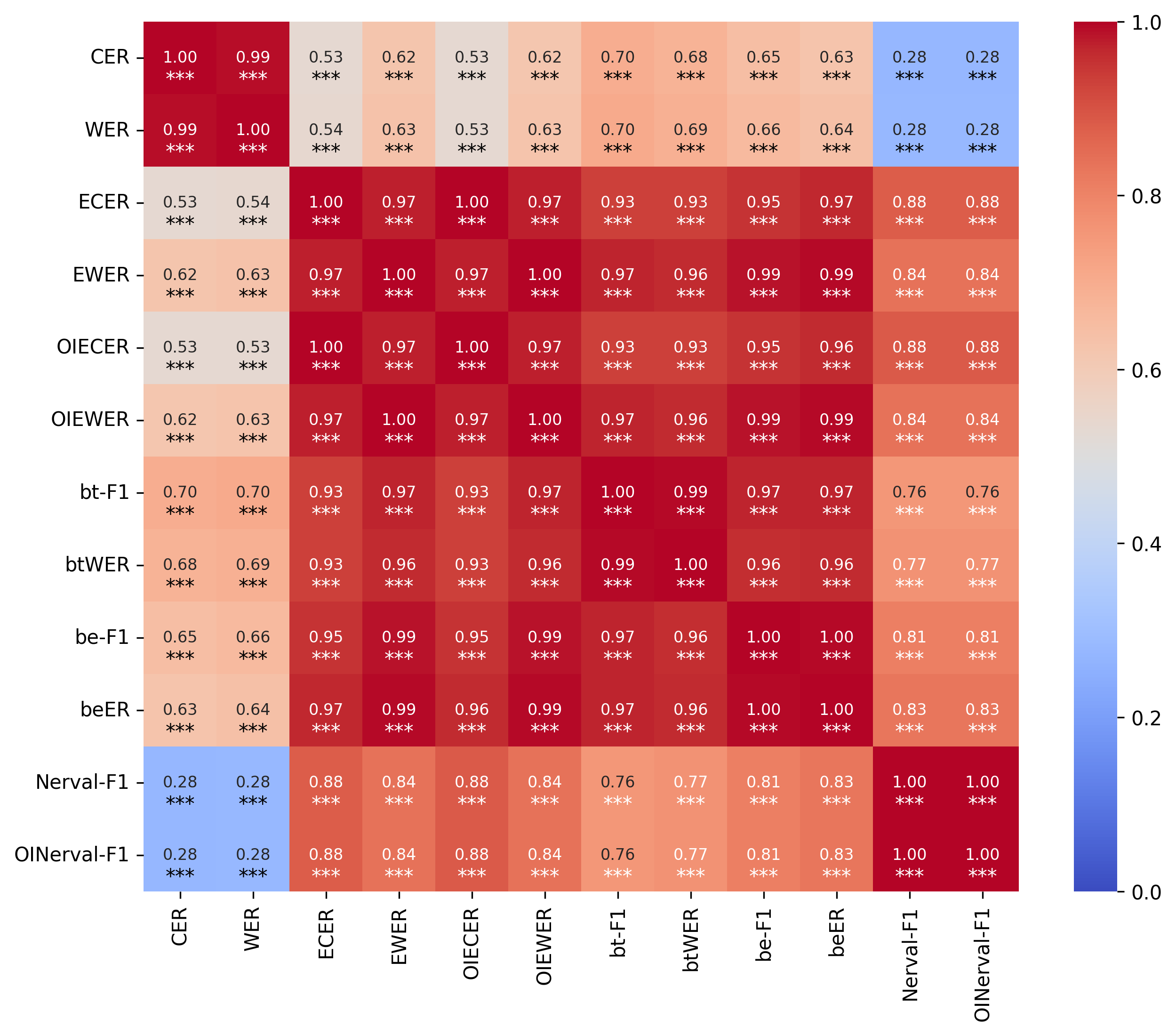}
         \caption{Esposalles \cite{Esposalles}}
         \label{fig:esposalles_s_corr}
     \end{subfigure}   
    \begin{subfigure}[b]{0.49\textwidth}
         \includegraphics[width=\textwidth, center]{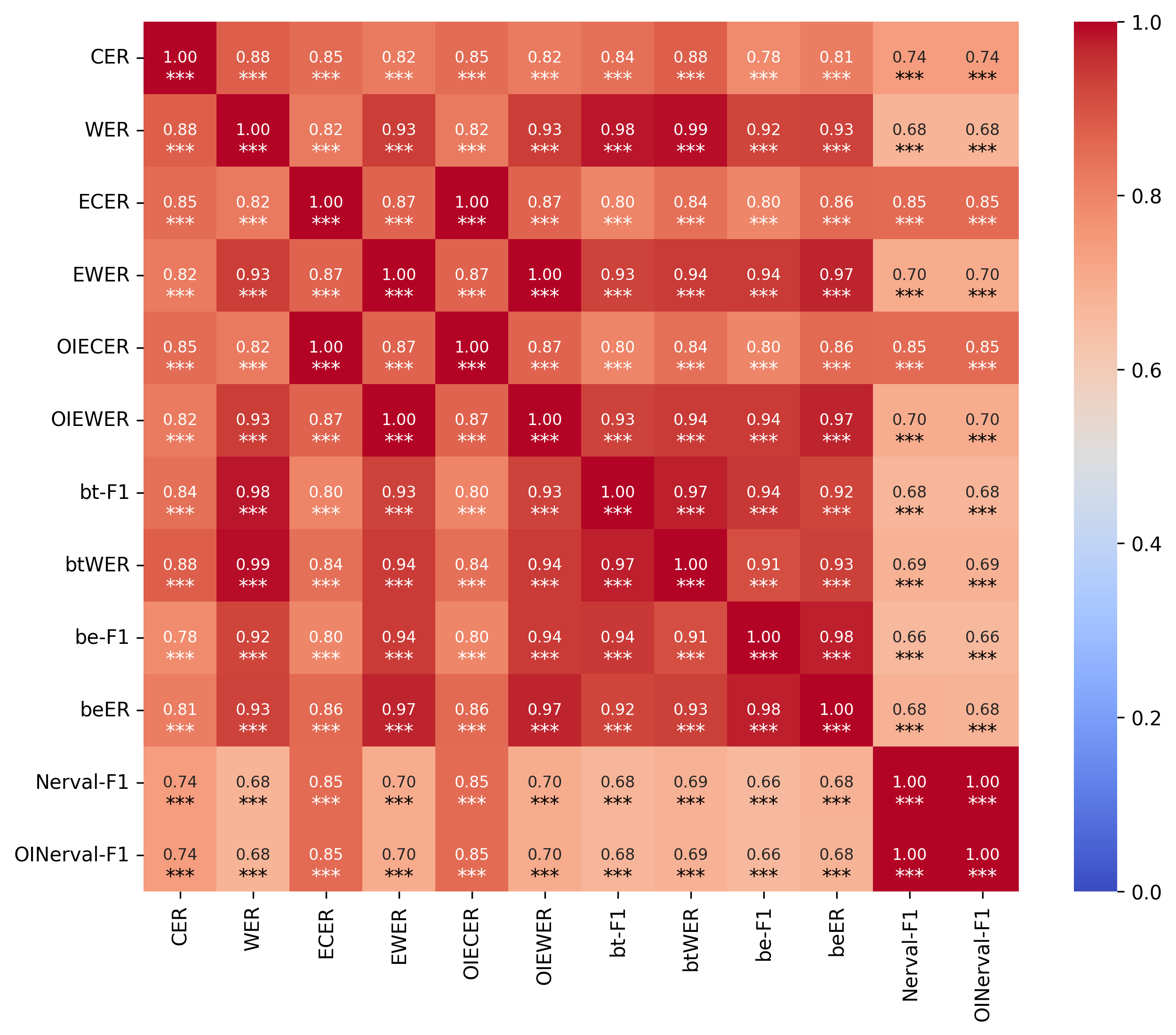}
         \caption{POPP \cite{POPP}}
         \label{fig:popp_s_corr}
     \end{subfigure}   
     \caption{Absolute rank correlation (Spearman) between the different metrics across the four datasets. The correlation value appears on each cell, as well as an indication of its p-value: \texttt{*} indicates a $p-value < 0.05$, \texttt{**} indicates a $p-value < 0.01$, and \texttt{***} indicates a $p-value < 0.001$. No star indicates that the correlation is not significant.}
    \label{fig:rank-correlation}
\end{figure}

We present the correlation matrix between different metrics on all four datasets. The linear correlation matrix is presented in Table \ref{fig:linear-correlation}, and the rank correlation matrix in Table \ref{fig:rank-correlation}. Note that we provide absolute correlations, as the sign of the correlation depends on the nature of the metrics, whether they are error rates (best at 0) or recognition rates (best at 100).

\subsection{Qualitative Analysis Of Different Metrics}
\label{sec:qualitative}
Finally, we present in Table \ref{tab:qualitative} a qualitative analysis of various metrics presented in this study across five prediction scenarios on Simara.

\begin{table}[!htb]
     \centering
     \caption{Qualitative analysis of metrics on specific examples.}
     \label{tab:qualitative}
     \begin{tabular}{p{0.1\textwidth}p{0.9\textwidth}}
     \toprule
          \textbf{Label} & \texttt{\color{yellow}<title>\color{black}COCATRIX (Jean-François) receveur du grenier à sel de Pithivers \color{pink}<analysis>\color{black}Contre Denis QUIROT \color{blue}<date>\color{black}26 mai 1770 \color{purple}<serie>\color{black}X1A \color{green}<article>\color{black}4678 \color{orange}<reference>\color{black}270} \\
          \midrule
          \textbf{Case 1} & \textbf{Almost perfect prediction.} As expected, all metrics yield very good scores when there are very few transcription and tagging errors.\\ 
          & \texttt{\color{yellow}<title>\color{black}COCATRIX (Jean-François) receveur du grenier à sel de Pithivers \color{pink}<analysis>\color{black}Contre Denis QUIROT \color{blue}<date>\color{black}26 mai 1770 \color{purple}<serie>\color{black}X1A \color{green}<article>\color{black}4678 \color{orange}<reference>\color{black}270} \\
          \midrule
          \textbf{Case 2} & \textbf{Missing words in a long entity} In this scenario, the text errors are concentrated in a single entity, which is consequently not matched. Entity-based metrics such as Nerval and the Bag-of-Entity F1 score are acceptable because only one entity is rejected. However, word-based metrics such as EWER and BoTW are low since many words are missed compared to the total number of words in the document. \\
          & \texttt{\color{yellow}<title>\color{black}COCATRIX (Jean-François) \color{pink}<analysis>\color{black}Contre Denis QUIROT \color{blue}<date>\color{black}26 mai 1770\color{purple}<serie>\color{black}X1A \color{green}<article>\color{black}4678 \color{orange}<reference>\color{black}270} \\
          \midrule
          \textbf{Case 3} & \textbf{Single word entity missing} In this scenario, a single entity (\texttt{reference}) consisting of a single word is missed. In this case, entity-based metrics such as Nerval and the Bag-of-Entity F1 score are acceptable because only one entity is rejected. Word-based metrics such as EWER and BoTW would be better or equal to entity-based metrics because all other words are correctly recognized. \\
          & \texttt{\color{yellow}<title>\color{black}COCATRIX (Jean-François) receveur du grenier à sel de Pithivers \color{pink}<analysis>\color{black}Contre Denis QUIROT \color{blue}<date>\color{black}26 mai 1770 \color{purple}<serie>\color{black}X1A \color{green}<article>\color{black}4678} \\
          \midrule
         \textbf{Case 4}  & \textbf{Uniform distribution of text errors}. In this scenario, the text errors are evenly distributed throughout the document. The Nerval F1 will be perfect because all entities will be matched (below the 30\% CER threshold). The ECER metric will be in an acceptable range. However, the word-based metrics (EWER, bt-F1, bt-WER) and the entity-based metrics (be-F1, be-WER) will be very low because many words contain at least one character error. \\
         & \texttt{\color{yellow}<title>\color{black}COCATRIX (Jean Francois) receveur du greniers à sel de Pithiver \color{pink}<analysis>\color{black}Contre Denis QUIROT \color{blue}<date>\color{black}26 mai 1771 \color{purple}<serie>\color{black}X1A- \color{green}<article>\color{black}4678 \color{orange}<reference>\color{black}270} \\
          \midrule 
          \textbf{Case 5} & \textbf{Tagging error}. In this scenario, there is no text error, but the model has swapped two entity tags (\texttt{analysis} and \texttt{title}). In this case, all metrics are bad.\\ 
          & \texttt{\color{pink}<analysis>\color{black}COCATRIX (Jean-François) receveur du grenier à sel de Pithivers \color{yellow}<title>\color{black}Contre Denis QUIROT \color{blue}<date>\color{black}26 mai 1770 \color{purple}<serie>\color{black}X1A \color{green}<article>\color{black}4678 \color{orange}<reference>\color{black}270} \\
    \bottomrule
    \end{tabular}
\end{table}